# CIS-Transolver: A Cross-layer Incremental-State Neural Operator for Fine-Scale Automotive Aerodynamic Flow Prediction

Jianglong Liu[a,b,c], Zhicheng Zhu[a,c], Jia Hao[a,b,c], Liangyue Jia[a,b,c,*] ,Wenxuan He[d]

[a] School of Mechanical Engineering, Beijing Institute of Technology, Beijing 100081, China

[b] Yangtze Delta Region Academy of Beijing Institute of Technology, Jiaxing 314019, China

[c] Key Laboratory of Industry Knowledge & Data Fusion Technology and Application，Ministry of Industry and Information Technology，Beijing Institute of Technology，Beijing 100081

[d] School of Big Data and Software Engineering, Chongqing University, Chongqing 400044, China

* Corresponding author: Liangyue Jia

E-mail address: jly1117@bit.edu.cn

## Abstract

Neural operators offer an efficient data-driven approach to predicting complex automotive aerodynamic flows. However, spectral bias limits their ability to reconstruct fine-scale flow variations around geometric features such as side mirrors and wheel arches. Reliable assessment of local geometric changes therefore requires models that capture fine spatial detail together with global flow interactions. Inspired by iterative refinement and state evolution, we propose CIS-Transolver, a cross-layer incremental-state neural operator that integrates local multiscale information into global flow modeling. The method extracts local feature variations and encodes them as zeroth- to second-order geometric moments. These encodings are combined with information from earlier layers to form a cross-layer memory. Memory-derived corrections enter the value branch of Physics-Attention, allowing local details to inform global physical-state interactions throughout the network. We systematically evaluate CIS-Transolver on three public automotive aerodynamics benchmarks, ShapeNet-Car, AhmedML, and DrivAerNet++, covering surface pressure fields and volumetric pressure and velocity fields. CIS-Transolver achieves the lowest mean relative L2 error among the evaluated methods for all five prediction metrics across the three benchmarks. Relative error reductions average 8.72% compared with the best baseline for each metric. Also achieves the lowest high-frequency spectral errors across all five metrics, with an average relative reduction of 16.06% compared with the best baseline. These results support cross-layer reuse of local information for fine-scale automotive flow prediction.

## Keywords

Automotive aerodynamics; Flow-field prediction; Neural operators; Multiscale modeling; Cross-layer memory

## 1. Introduction

Local changes in vehicle geometry can substantially alter the surface pressure distribution and downstream wake structure, so each change to the vehicle body shape during aerodynamic design and optimization may require a new assessment of the flow response. Fast and accurate prediction of external vehicle flows within the time available for design iterations is therefore essential for efficient aerodynamic development. High-fidelity computational fluid dynamics (CFD) remains a key approach for computing flow fields around vehicle bodies. However, repeated mesh generation, numerical discretization, and iterative solution for different geometries incur substantial computational costs, limiting its use in rapid design [1,2]. As such, data-driven methods increasingly support rapid flow-field prediction. Neural operators learn mappings between function spaces and can directly map vehicle geometry and operating conditions to aerodynamic flow fields, providing efficient surrogates for costly CFD simulations [3–5]. Advances in Fourier neural operators, Transformers, and graph neural operators improve the treatment of complex boundaries, irregular discretizations, and large computational domains, enabling rapid prediction of surface pressure fields and volumetric velocity and pressure fields [6–8]. Transolver++ [9] and AB-UPT [10] further extend these advances to high-resolution automotive aerodynamics.

A major challenge for neural operators is their limited accuracy in predicting local flow details [11–14]. Spectral bias contributes to this difficulty, which leads models to prioritize low-frequency components that dominate the energy content and vary smoothly in space, while underfitting rapidly varying local components and producing overly smooth predictions. In external vehicle flows, even when a model achieves low global prediction error, excessive smoothing of local flow structures can obscure aerodynamic differences between designs and reduce the usefulness of predictions for design assessment and optimization.

To improve the representation of spatial variations, prior studies have explored multiscale and convolutional operator architectures [15–17], local spatial interaction modeling [18–20], and derivative-based training objectives [21,22]. These approaches offer complementary means of capturing local features and interactions across scales. However, accurately reconstructing fine-scale flow structures while modeling global flow interactions remains challenging.

Iterative refinement and state evolution offer further opportunities to improve local fine-scale flow-field prediction. PDE-Refiner progressively corrects predicted fields through multistep denoising, improving the recovery of low-amplitude spatial frequency components [23]. CoEvol-NO adopts an error-driven predictor–corrector scheme to jointly update persistent latent states and mesh representations, combining state evolution with pointwise feedback [24]. These advances motivate us to integrate local multiscale modeling with cross-layer state updates to improve the reconstruction of fine-scale variations in complex automotive aerodynamic flows.

As such, we propose CIS-Transolver, a cross-layer incremental-state neural operator for fine-scale automotive aerodynamic flow prediction. Building on the Physics-Attention mechanism of Transolver, CIS-Transolver integrates local multiscale analysis with cross-layer state updates. The model extracts local feature variations at different spatial scales and encodes them as geometric moments to construct incremental states that persist across operator layers. At selected layers, the memory is updated by combining current local observations with historical states. At each layer, a memory-conditioned readout generates feature corrections that are incorporated into the Value branch of Physics-Attention. This design allows fine-scale information to contribute to the modeling of global physical interactions throughout successive operator layers.

We evaluate CIS-Transolver on three public automotive aerodynamics benchmarks, AhmedML [25], DrivAerNet++ [26], and ShapeNet-Car [27], covering surface pressure prediction and volumetric pressure and velocity prediction. We compare the proposed method with representative neural operators, including Transolver and CoEvol-NO. CIS-Transolver achieves the lowest mean relative $L_2$ error across all five prediction metrics, with an average error reduction of 8.72% relative to the best baseline for each metric. On the DrivAerNet++ test set, CIS-Transolver reduces pressure and velocity field-increment errors by 8.3% and 2.6%, respectively, relative to Transolver in the smallest evaluated separation bin. Mean pressure and velocity spectral errors over the high-frequency range $13 \leq k \leq 24$ decrease by 6.02% and 3.04%, respectively, relative to the same baseline. Ablations on DrivAerNet++ show that the full model reduces pressure and velocity prediction errors by approximately 3.7% and 4.2%, respectively, relative to the variant without the CIS branch. The variant without cross-layer state reuse retains local multiscale analysis, moment encoding, and current-layer Value feedback. Relative to this variant, the full model reduces high-frequency pressure spectral error by 3.54%. The main contributions of this work are summarized as follows:

**1)** We develop CIS-Transolver, a neural operator that combines local fine-scale modeling with cross-layer incremental-state updates. The method encodes multiscale local feature variations as geometric moments and integrates historical state information into Physics-Attention through Value feedback. This enables local flow information to contribute to global physical interaction modeling across successive operator layers.

**2)** We conduct a systematic evaluation of CIS-Transolver on three public automotive aerodynamics benchmarks covering different vehicle geometries and both surface and volumetric flow prediction tasks. The proposed method achieves the lowest prediction errors across all five evaluation metrics and improves the reconstruction of local flow structures and high-frequency spatial components.

## 2. Related Work

### 2.1 Automotive aerodynamic surrogate modeling and geometric representations

Drag prediction is a common objective of automotive aerodynamic surrogate modeling. Models use geometric parameters and learned low-dimensional representations [28-31], rendered views [32], or three-dimensional surface geometry [33]; wake-field information can also supplement limited drag data [34]. These approaches enable rapid assessment of integral aerodynamic performance. Their principal aerodynamic target, however, is an integrated response rather than a three-dimensional pressure or velocity field. Estimating this response alone does not reveal local flow distributions or establish whether individual flow structures are accurately represented.

Full-field prediction requires representations that relate vehicle geometry to spatially distributed flow variables. Common-domain mappings and reduced-order reconstruction establish such correspondences [27,35]. Convolutional surrogates use signed-distance fields to estimate drag coefficients and, in some settings, velocity fields [36], whereas other CNN-based methods predict pressure and velocity fields around vehicles [37–38]. The applicability of these approaches depends on how geometry and flow are represented. Parametric descriptions restrict shape variation to predefined variables, reduced-order models reconstruct fields within a learned basis, and grid-based methods require a prescribed spatial layout. These constraints can limit the transfer of a trained model to new geometric representations or discretizations.

## 2.2 Neural operators for complex vehicle geometries

Neural operators learn mappings between function spaces [3–5]. Deep operator networks and Fourier neural operators provide representative formulations based on coordinate-conditioned prediction and spectral integral kernels, respectively. For vehicle aerodynamics, subsequent developments address two coupled concerns: accommodating irregular geometries and modeling spatial interactions at practical computational cost. Geometry-aware Fourier models connect irregular domains to regular latent representations [6,39], while graph-based surrogates model local interactions on unstructured meshes [40–42]. This retains neighborhood relationships around complex boundaries without first mapping the physical domain onto a regular grid. Attention-based formulations accommodate heterogeneous inputs and irregular sampling [43], and compact latent representations reduce the cost of repeated global interactions [44,45]. These routes improve geometric flexibility and computational scalability through different choices of spatial representation and interaction.

Transolver models long-range physical correlations by aggregating points into learned physical states and applying Physics-Attention within that state space [7]. With a fixed number of states, its computational cost grows linearly with the input point count. Automotive models further combine local and global interactions [8] and extend prediction to million-point geometries and high-resolution surface and volume fields [9,10]. More recent work addresses non-watertight geometries and adaptation across flow conditions [46,47], while repeated geometry conditioning and coarse-to-fine refinement connect global shape information to local field reconstruction [48,49]. Despite improvements in overall prediction, local high-gradient and fine-scale structures can remain under-resolved, partly because spectral bias favors low-frequency components during learning [11,14,21].

## 2.3 Fine-scale flow representation and refinement

Fine-scale prediction depends in part on the ability of a model to represent local spatial variations. Fourier feature mappings and periodic activations improve the approximation of high-frequency functions in coordinate networks [50,51]. Wavelet-based operators provide spatially localized multiresolution representations [52,53]. Extensions using U-Net paths or multiscale Fourier branches further improve the approximation of local and oscillatory components [54,55]. These methods address complementary properties of fine-scale signals: their frequency content and, for wavelet representations, their spatial localization.

Local interactions and gradient-based supervision provide complementary means of improving fine-scale accuracy. The spatially aware transformer operator (SATO) combines global physics attention with local attention for vehicle surface-pressure prediction [20]. The hierarchical attention neural operator (HANO) models interactions across scales and uses a loss that penalizes spatial-derivative errors to mitigate spectral bias [21]. Neighborhood interactions encode relationships among nearby samples, while derivative-based losses emphasize errors in local field variations. A small pointwise field error can coexist with substantial gradient error, making derivative supervision particularly relevant to sharp spatial variations. Together, these mechanisms improve how local structure is captured and fitted, complementing changes to the representation basis.

Iterative refinement addresses errors by repeatedly updating predictions or internal representations. PDE-Refiner uses multistep denoising to better capture low-amplitude spatial-frequency components and improve long-term prediction [23]. Generative approaches extend fine-scale reconstruction through flow super-resolution and conditional field generation [56,57]. At the representation level, CoEvol-NO jointly updates persistent latent states and mesh representations through an error-driven predictor–corrector mechanism [24]. Although PDE-Refiner and CoEvol-NO update predicted fields and latent representations, respectively, both demonstrate the potential of progressive updates to improve prediction accuracy. Inspired by these studies, we propose CIS-Transolver, which couples local multiscale analysis with cross-layer incremental state updates to incorporate fine-scale information into Physics-Attention and improve both local and overall flow-field accuracy.

## 3. CIS-Transolver

CIS-Transolver combines a geometry-aware encoder, stacked CIS operator layers, and a field decoder (Figure 1a). The encoder maps coordinates, geometric attributes, and case-level conditions to initial hidden features. Each CIS layer couples global interactions among learned physical states with local feature analysis and cross-layer memory (Figure 1b). At designated refresh layers, multiscale analysis extracts mid-scale and fine-scale variations on shared geometric support.

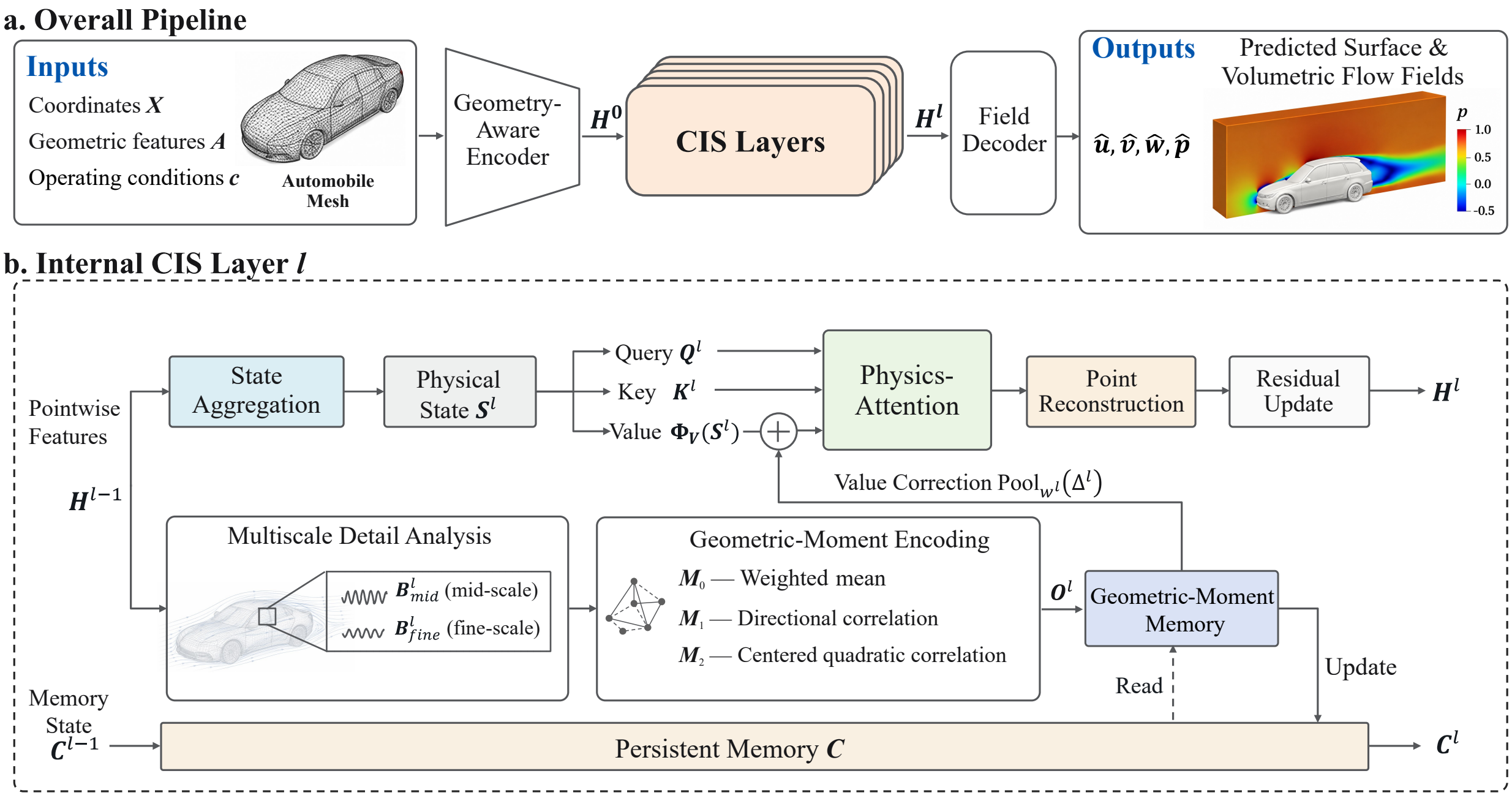

Figure 1. Overall architecture of CIS-Transolver. (a) Geometry-aware encoding, stacked CIS layers, and field decoding. (b) A CIS layer couples Physics-Attention with multiscale analysis, geometric-moment encoding, and memory fusion. Memory is refreshed at layers 1, 4, and 8 and read at every layer to generate corrections for the attention Value branch.

## 3.1 Problem formulation and geometry-aware feature encoding

Let $X \in \mathbb{R}^{N\times 3}$ denote the three-dimensional coordinates of $N$ sampling points, $A$ the corresponding pointwise attributes, and $c$ the case-level geometric descriptors and operating conditions. CIS-Transolver learns a mapping from $(X, A, c)$ to the predicted field $\hat{Y}$. For the surface task, the model predicts pressure at each point; for the volumetric task, it jointly predicts three velocity components and pressure.

The encoder maps pointwise attributes, explicit geometric features, and case-level conditions through separate branches and fuses them into the initial hidden representation $H^0$. Explicit geometric features comprise flow-aligned coordinates, a near-wall distance response, and a downstream position response. Availability masks control the corresponding encoding branches when geometric attributes are missing. Dataset-specific inputs are mapped to a common 256-dimensional hidden representation. The input fields and branch configurations are given in Appendix B.1.

## 3.2 Geometry-based multiscale analysis

We extract local feature differences through geometry-based smoothing and reconstruction on a shared point–anchor support. The distance-weighted connections are fixed within each forward pass, so observations at different depths use the same spatial neighborhoods. Let $P_g \in \mathbb{R}^{N \times M_g}$ denote the sparse anchor-to-point interpolation matrix for $N$ sample points and $M_g$ anchors. The interpolation weights depend on distances between points and their neighboring anchors and are normalized for each point. The point-to-anchor aggregation and anchor-domain smoothing operators are defined as

$$D_g = diag\left(P_g^T 1_N\right), \quad R_g = D_g^{-1} P_g^T, \quad K_g = R_g P_g \tag{1}$$

Here, $D_g$ is the diagonal matrix of total weights received by the anchors, $R_g$ aggregates pointwise features at the anchors, and $K_g$ smooths anchor-domain features by prolongation to the point domain followed by aggregation back to the anchors. In Eq. (1), $1_N$ is an all-ones vector, $diag$ constructs a diagonal matrix, and superscript $T$ denotes transposition. We use 64 geometric anchors, with each point connected to its three nearest anchors. Dataset-specific anchor selection and interpolation weights are given in Appendix B.1.

At a memory-refresh layer $l$, layer normalization and a head-specific channel projection map $H^{l-1}$ to $F^l$ (Appendix B.1). The shared geometric support gives the mid-scale and fine-scale observations:

$$\begin{gathered} B_{mid}^l = P_g\left(I_{M_g} - K_g\right) R_g F^l \\ B_{fine}^l = F^l - L_g\left(F^l\right) \end{gathered} \tag{2}$$

Here, $B_{mid}^l$ and $B_{fine}^l$ denote the mid-scale and fine-scale observations, respectively. The mid-scale observation captures differences between anchor-averaged features and their further-smoothed counterparts, mapped back to the sampling points. The fine-scale observation captures pointwise deviations from the local reconstruction. Their spatial scales are determined by the anchor distribution and interpolation weights. $I_{M_g}$ is the identity matrix, and $L_g$ is the dataset-specific local reconstruction operator listed in Table B.1 (Appendix B.1).

### 3.3 Cross-layer geometric moment memory

CIS-Transolver encodes the mid-scale and fine-scale observations $B_{mid}^l$ and $B_{fine}^l$ from Eq. (2) in a geometric moment memory. Moments of orders zero to two capture local amplitudes and spatial structure. As shown in Figure 2, gated fusion updates the memory at selected layers, while every layer reads it to correct the attention values.

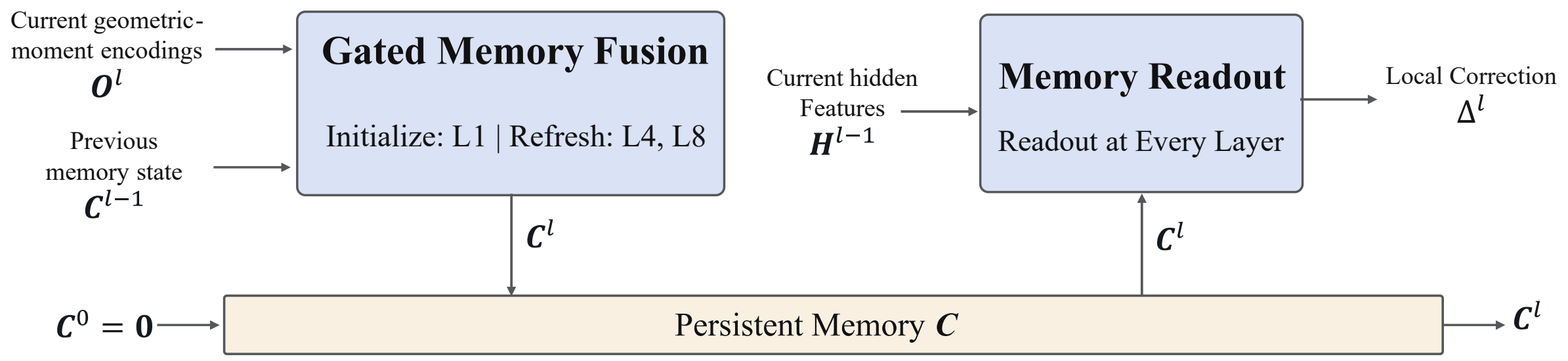

Figure 2. Geometric-moment memory. Gated fusion refreshes memory at layers 1, 4, and 8 and retains it elsewhere. Layerwise readout combines memory and hidden features to generate a Value correction.

For anchor $a$, the normalized weights in row $a$ of $R_g$ give centered and scaled coordinates $\xi_{ia}$ (Appendix B.2). The geometric bases are

$$\varphi_{ia}^{(0)}=1, \varphi_{ia}^{(1)}=\xi_{ia}, \\ \varphi_{ia}^{(2)}=\mathrm{vec}_{sym}\left(\xi_{ia}\xi_{ia}^{T}-\Sigma_a\right). \tag{3}$$

Here, $\Sigma_a$ is the weighted mean of $\xi_{ia}\xi_{ia}^{T}$, and $\mathrm{vec}_{sym}$ extracts the six independent symmetric entries. The basis dimensions are $d_0=1$, $d_1=3$ and $d_2=6$. These bases are shared across layers.

For point $i$ in each head, let $b_{i,mid}^{l}=\left(B_{mid}^{l}\right)_{i,:}^{T}$ and $b_{i,fine}^{l}=\left(B_{fine}^{l}\right)_{i,:}^{T}$ denote the observations with $r$ channels. For $s\in\{mid, fine\}$, the moments and encodings are

$$M_{a,s}^{l,(k)}=\sum_{i=1}^{N}\left(R_g\right)_{ai}\varphi_{ia}^{(k)}\left(b_{i,s}^{l}\right)^{T}\in\mathbb{R}^{d_k\times r}, \\ o_{a,s}^{l,(k)}=E_k^l\left(\mathrm{vec}\left(M_{a,s}^{l,(k)}\right)\right)\in\mathbb{R}^{r}, k=0,1,2. \tag{4}$$

The three orders describe mean amplitudes, directional correlations, and correlations with quadratic coordinate terms. Here, $\mathrm{vec}$ vectorizes each moment, and $E_k^l$ maps it to $r$ channels. Concatenating the encodings gives $o_{a,s}^{l}\in\mathbb{R}^{3r}$, with separate memories for the two scales.

At refresh layers, order-specific linear maps produce history encodings, which combine with the current encodings to determine the update gates:

$$h_{a,s}^{l,(k)}=T_k^l\left(c_{a,s}^{l-1,(k)}\right), \\ g_{a,s}^{l}=\sigma\left(G^l\left(\left[\ln\left(o_{a,s}^{l}\right),\ln\left(h_{a,s}^{l}\right)\right]\right)\right). \tag{5}$$

Here, $h_{a,s}^{l}$ concatenates the history encodings. Layer normalization acts over $3r$ channels, brackets denote concatenation, and $\sigma$ is the sigmoid. The affine map $G^l$ produces one scalar gate per anchor, scale, and order, shared across its $r$ channels. The memory update is

$$c_{a,s}^{l,(k)} = \begin{cases} \left(1 - g_{a,s}^{l,(k)}\right) h_{a,s}^{l,(k)} + g_{a,s}^{l,(k)} o_{a,s}^{l,(k)}, & l \in U, \\ c_{a,s}^{l-1,(k)}, & l \notin U. \end{cases} \tag{6}$$

The components $c_{a,s}^{l,(k)}$ form $C^l$, initialized as $C^0 = 0$. The eight-layer backbone uses $U = \{1,4,8\}$, based on the sensitivity study in Appendix A.3. Updates precede attention, and Section 3.4 describes the Value readout.

### 3.4 Memory-conditioned physics attention

Moment memory conditions the values exchanged between physical states. At each layer, the current hidden features determine the state assignments and query–key interactions, while memory readout supplies an additive correction to the state values. Memory is read at every layer, including layers where it is retained without refresh.

At layer $l$, the mid-scale and fine-scale components of $C^l$ jointly determine anchor-level modulation $A^l$ and bias $B^l$. These parameters condition the readout of the current hidden representation $H^{l-1}$ to produce a pointwise correction $\Delta^l$:

$$\left(A^l, B^l\right) = \Psi^l\left(C^l\right), \Delta^l = F_{read}^l\left(H^{l-1}; A^l, B^l\right). \tag{7}$$

The readout $F_{read}^l$ interpolates the anchor parameters to points and applies multiplicative modulation and additive correction to a low-rank feature projection. Channel mapping, amplitude matching, and gating yield the final correction. At refresh layers, the bias is filtered by $I_{M_g} - K_g$ before interpolation. The readout operations are specified in Appendix B.2.

For each attention head, a learned affine map of the normalized layer input produces pointwise values $V_{pt}^l$ and state-assignment logits. A temperature-scaled softmax over states gives $W^l$. Averaging the pointwise values with mass-normalized assignment weights yields state features $S^l$. Memory then corrects the state values:

$$V^l = S^l U_V^l + \left(\Lambda^l\right)^{-1} \left(W^l\right)^T \Delta^l. \tag{8}$$

Here, $\Lambda^l$ is the diagonal matrix of state assignment masses, bounded below by a positive constant $\varepsilon$, and $U_V^l$ is a learned head-specific linear projection. State attention uses linear query and key projections of $S^l$ to mix the corrected values. Its outputs return to points through $W^l$, followed by head concatenation, output projection, residual updates, and the feed-forward block. A final decoder maps the hidden representation to the predicted field.

Training uses the final prediction and reference field in standardized target units. Let $e = \hat{Y} - Y$ denote their difference, and define the geometric detail operator $H_g = I_N - P_g R_g$. The field and detail losses are $L_{field} = E(e)$ and $L_{detail} = E(H_g e)$, respectively. The training objective is

$$L_{train} = L_{field} + \lambda_d L_{detail} \tag{9}$$

The detail-loss weight $\lambda_d$ is fixed at 0.1 for all three datasets. The operator $E$ averages squared errors over valid entries, weighting the velocity and pressure groups equally for volumetric tasks and using pressure alone for surface prediction. The detail term emphasizes errors relative to geometry-based local averages. Both terms supervise the final output. Periodic detail-loss evaluations are multiplied by their interval. Inference uses the same feature-based memory updates and Value readout, followed by the final field decoder.

# 4 Experimental results and analysis

## 4.1 Datasets and experimental setup

We evaluate surface-pressure prediction on ShapeNet-Car and volumetric pressure and three-component velocity prediction on AhmedML and DrivAerNet++. The volumetric tasks use fixed case subsets to accommodate data-storage requirements and training costs. Each task is split by complete geometry into training, validation, and test sets. Training and evaluation use the valid points retained after preprocessing: 3,586 per ShapeNet-Car case and approximately 32,000 per volumetric case. Appendix A reports the case splits and counts, input features, and preprocessing rules.

The comparison includes MeshGraphNet[40], Geo-FNO[39], GINO[6], Transolver[7], Transolver++[9], AeroGTO[8], SATO[20], AB-UPT[10], and CoEvol-NO[24]. All models are evaluated on the same test cases for each task, retaining their respective input encodings and training configurations. Appendix B describes the implementation, training objectives, and model-selection settings of CIS-Transolver.

Quantitative evaluation uses the discrete relative $L_2$ error, $\mathtt{rel}_q = \frac{\|\hat{q} - q\|_2}{\|q\|_2}$, where $q$ and $\hat{q}$ denote the reference and predicted field vectors, respectively. Each case is evaluated on all valid points, with the three velocity components concatenated to obtain the velocity error. Pressure and velocity errors are reported separately and averaged equally over test geometries.

### 4.2 Predictive performance on benchmark datasets

Table 1 reports mean errors across the three benchmarks. CIS-Transolver achieves the lowest error for all five prediction metrics, with an average reduction of 8.72% relative to the best baseline for each metric.

Table 1 Mean relative $L_2$ errors on the three automotive aerodynamics benchmarks.

| **Method** | **ShapeNet-Car** | **AhmedML** | | **DrivAerNet++** | |
|---|---|---|---|---|---|
| | $\mathrm{rel}_p \downarrow$ | $\mathrm{rel}_p \downarrow$ | $\mathrm{rel}_U \downarrow$ | $\mathrm{rel}_p \downarrow$ | $\mathrm{rel}_U \downarrow$ |
| MeshGraphNet | 0.2829 | 0.1529 | 0.0972 | 0.1693 | 0.1722 |
| Geo-FNO | 0.3856 | 0.1160 | 0.0965 | 0.2003 | 0.1994 |
| GINO | 0.5130 | 0.1593 | 0.1273 | 0.2999 | 0.2519 |
| Transolver | <u>0.1490</u> | 0.0812 | 0.0766 | 0.1501 | 0.1428 |
| Transolver++ | 0.1875 | 0.0803 | 0.0761 | 0.1493 | 0.1459 |
| AeroGTO | 0.2213 | 0.1087 | 0.0868 | 0.1517 | 0.1671 |
| SATO | 0.2158 | 0.0820 | 0.0772 | 0.1531 | 0.1427 |
| AB-UPT | 0.1584 | 0.1131 | 0.0774 | 0.2072 | 0.1793 |
| CoEvol-NO | 0.1571 | <u>0.0796</u> | <u>0.0750</u> | <u>0.1449</u> | <u>0.1421</u> |
| **CIS-Transolver(Ours)** | **0.1246** | **0.0748** | **0.0633** | **0.1405** | **0.1385** |

Note: $\mathrm{rel}_p$ and $\mathrm{rel}_U$ denote the relative $L_2$ errors in pressure and three-component velocity, respectively. The downward arrow (↓) indicates that lower values are better. Results are averaged equally over test geometries. ShapeNet-Car is evaluated for surface pressure only. Bold and underlined values indicate the lowest and second-lowest errors in each column, respectively.

Relative to the best baseline for each metric, CIS-Transolver reduces the surface-pressure error on ShapeNet-Car by 16.4%. On AhmedML, pressure and velocity errors decrease by 6.0% and 15.6%, respectively; the corresponding reductions on DrivAerNet++ are 3.0% and 2.5%. The largest gains occur in surface-pressure prediction on ShapeNet-Car and volumetric velocity prediction on AhmedML.

Table 2 reports the mean high-frequency spectral errors across the three benchmarks. CIS-Transolver achieves the lowest errors for all five prediction metrics, with an average reduction of 16.06% relative to the best baseline for each metric. The reductions reach 14.31% for ShapeNet-Car pressure prediction, 19.79% and 41.46% for AhmedML pressure and velocity prediction, and 3.28% and 1.45% for DrivAerNet++ pressure and velocity prediction, respectively. These results demonstrate that CIS-Transolver improves high-frequency flow-field reconstruction across the evaluated benchmarks, complementing its improvements in overall prediction accuracy.

Table 2 Mean high-frequency spectral errors on the three automotive aerodynamics benchmarks.

| **Method** | **ShapeNet-Car** | **AhmedML** | | **DrivAerNet++** | |
|---|---|---|---|---|---|
| | Spectral $p\downarrow$ | Spectral $p\downarrow$ | Spectral $u\downarrow$ | Spectral $p\downarrow$ | Spectral $u\downarrow$ |
| MeshGraphNet | 0.3291 | 0.0937 | 0.0734 | 0.3489 | 0.3238 |
| Geo-FNO | 0.4403 | 0.0725 | 0.1142 | 0.3702 | 0.3460 |
| GINO | 0.5850 | 0.1055 | 0.1200 | 0.5001 | 0.4449 |
| Transolver | <u>0.1775</u> | <u>0.0374</u> | 0.0571 | 0.3149 | 0.2743 |
| Transolver++ | 0.2219 | 0.0385 | 0.0554 | 0.3093 | 0.2796 |
| AeroGTO | 0.2614 | 0.0728 | 0.0831 | 0.3176 | 0.2994 |
| SATO | 0.2501 | 0.0384 | 0.0588 | 0.3131 | 0.2747 |
| AB-UPT | 0.1819 | 0.0876 | 0.0904 | 0.3586 | 0.3087 |
| CoEvol-NO | 0.1837 | 0.0393 | <u>0.0521</u> | <u>0.3046</u> | <u>0.2688</u> |
| **CIS-Transolver(Ours)** | **0.1521** | **0.0300** | **0.0305** | **0.2946** | **0.2649** |

Note: For each test case, the high-frequency relative spectral error is computed as the square root of the ratio of summed error power to summed reference power over ($13 \le k \le 24$), where $k$ denotes dimensionless spatial frequency. Results are then averaged equally across test geometries. Velocity errors include all three components. Evaluation uses all 3,586 surface points per ShapeNet-Car case and 4,096 sampled volume points per AhmedML and DrivAerNet++ case. Bold values indicate the lowest error in each column; ↓ indicates that lower values are better.

Figure 3 compares the errors on individual DrivAerNet++ test geometries. Each marker pairs the error of CIS-Transolver with that of a baseline on the same geometry, showing how the test-set improvements are distributed across cases.

CIS-Transolver has lower pressure and velocity errors than Transolver on all ten test geometries. Its velocity error is also lower than that of CoEvol-NO throughout the test set while its pressure error is lower on most geometries.

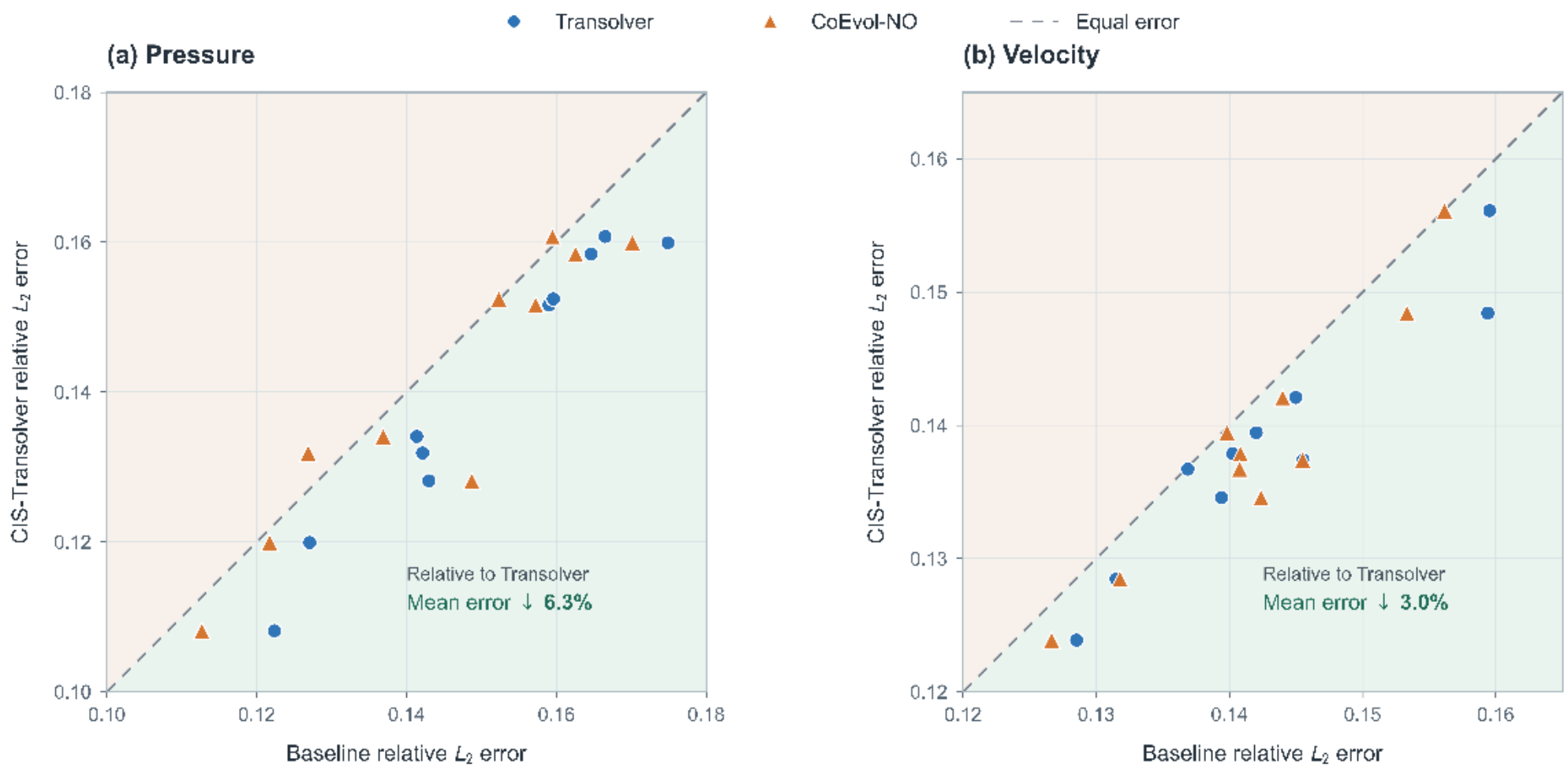

Figure 3 Case-wise paired relative $L_2$ errors on DrivAerNet++. Panels show (a) pressure and (b) three-component velocity. Each marker compares CIS-Transolver with one baseline on the same test geometry. The dashed lines indicate equal errors; markers below them denote lower errors for CIS-Transolver. All ten shared test geometries are included.

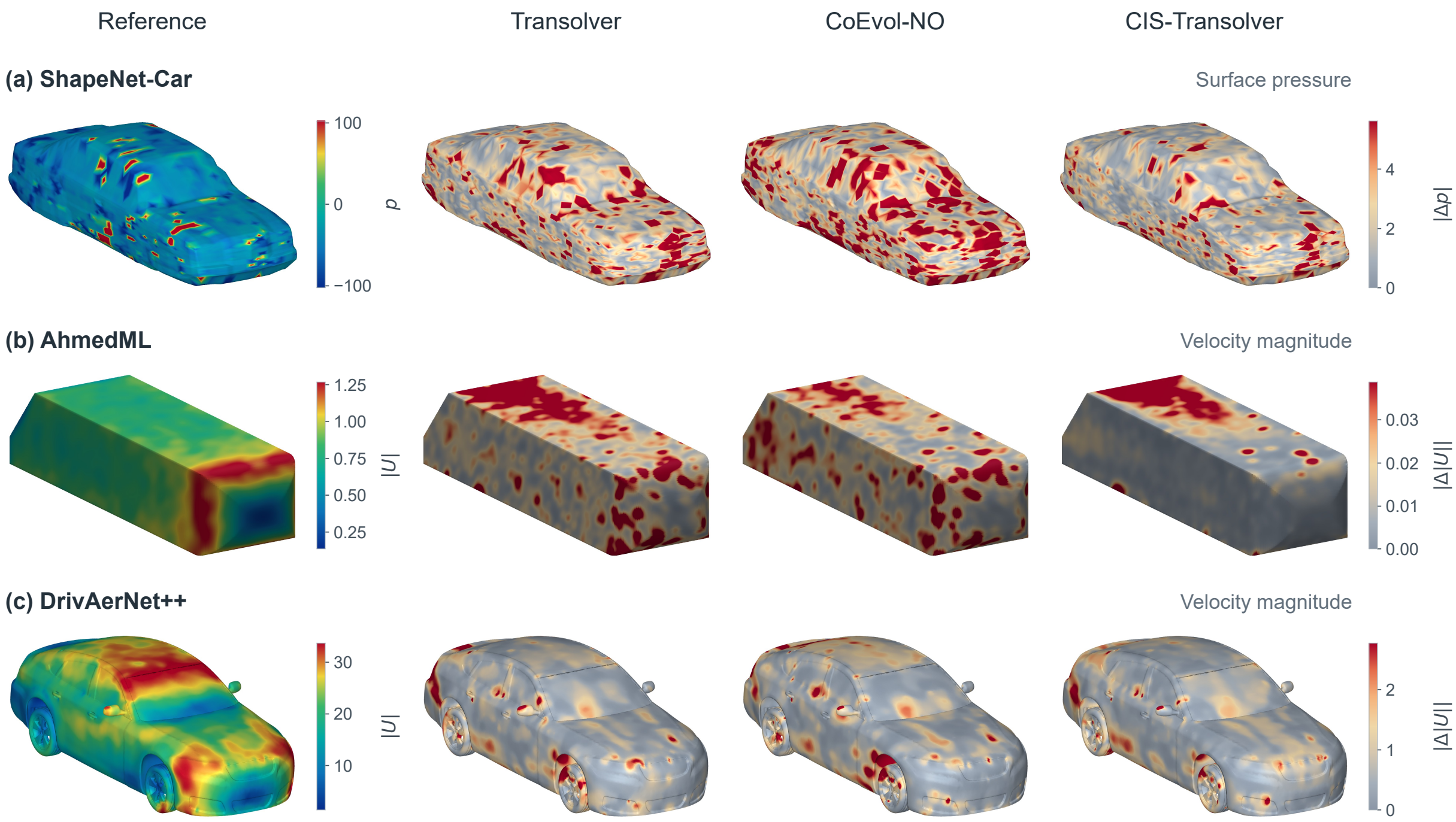

Figure 4 Reference fields and absolute prediction errors for three test geometries: (a) surface pressure on ShapeNet-Car; (b) velocity magnitude on AhmedML; and (c) velocity magnitude on DrivAerNet++. The first column shows the reference fields, followed by the errors of Transolver, CoEvol-NO, and CIS-Transolver. Error maps share a common color scale within each row.

Compared with Transolver and CoEvol-NO, CIS-Transolver reduces the extent of high-error pressure patches over the roof and side panels of the ShapeNet-Car geometry (Fig. 4(a)). On AhmedML, it lowers velocity-magnitude errors along the visible vertical faces (Fig. 4(b)). On DrivAerNet++, it reduces high-error regions around the front wheel arch (Fig. 4(c)).

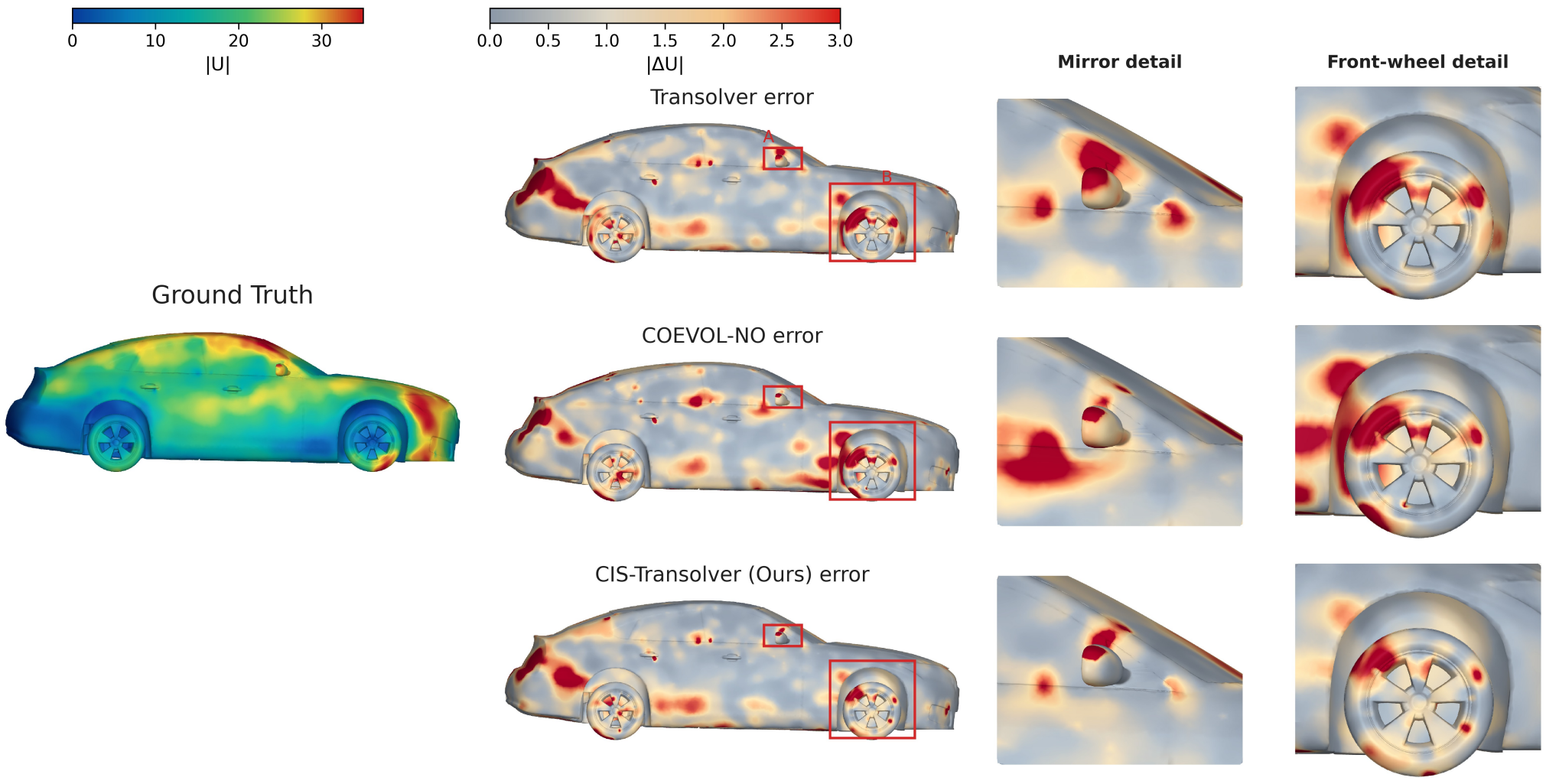

Figure 5 Velocity magnitude and local prediction errors for a DrivAerNet++ test vehicle. The reference field is shown on the left. The center column shows the errors of Transolver, CoEvol-NO, and CIS-Transolver, and the two rightmost columns show enlarged views of the mirror and front-wheel regions. Viewpoints, local regions, and error color scales are shared across models.

Figure 5 examines velocity reconstruction near the mirror and front wheel of a DrivAerNet++ test vehicle. Enlarged views resolve local differences that are less apparent in the full-field comparison. All models use the same viewpoint, local regions, and error color scale.

Near the mirror, Transolver shows strong error patches above the mirror and on the adjacent body, while CoEvol-NO has an extended high-error region beside and below it. CIS-Transolver reduces these patches and confines high errors to smaller regions. Near the front wheel, both baselines show relatively continuous high errors along the upper wheel edge and adjacent body; CIS-Transolver reduces their extent and weakens nearby error patches.

We extend the local comparisons in Figure 5 to all ten DrivAerNet++ test geometries by evaluating how accurately the models recover pressure and velocity differences between volume points at different separations (Figure 6).

All valid volume-point pairs are grouped into ten logarithmic bins over $0.02 < r^* \leq 0.25$, where $r^* = r/L$ is three-dimensional point separation normalized by vehicle length. The RMS error in predicted field increments is divided by the RMS reference increment, using common density-compensated weights and all three velocity components. Errors are computed per case and averaged equally over the ten geometries. CIS-Transolver yields the lowest mean pressure and velocity increment errors across all ten separation bins. In the shortest-separation bin, the errors are 8.3% and 2.6% lower than those of Transolver, respectively. With the lower limit unchanged, CIS-Transolver also retains the lowest pressure and velocity increment errors when all point pairs are pooled over windows with upper limits of $0.15\,L$, $0.20\,L$, or $0.30\,L$.

Overall, CIS-Transolver reduces full-field prediction errors, lowers local errors near the examined geometric features, and more accurately reconstructs pressure and velocity increments across the evaluated spatial separations.

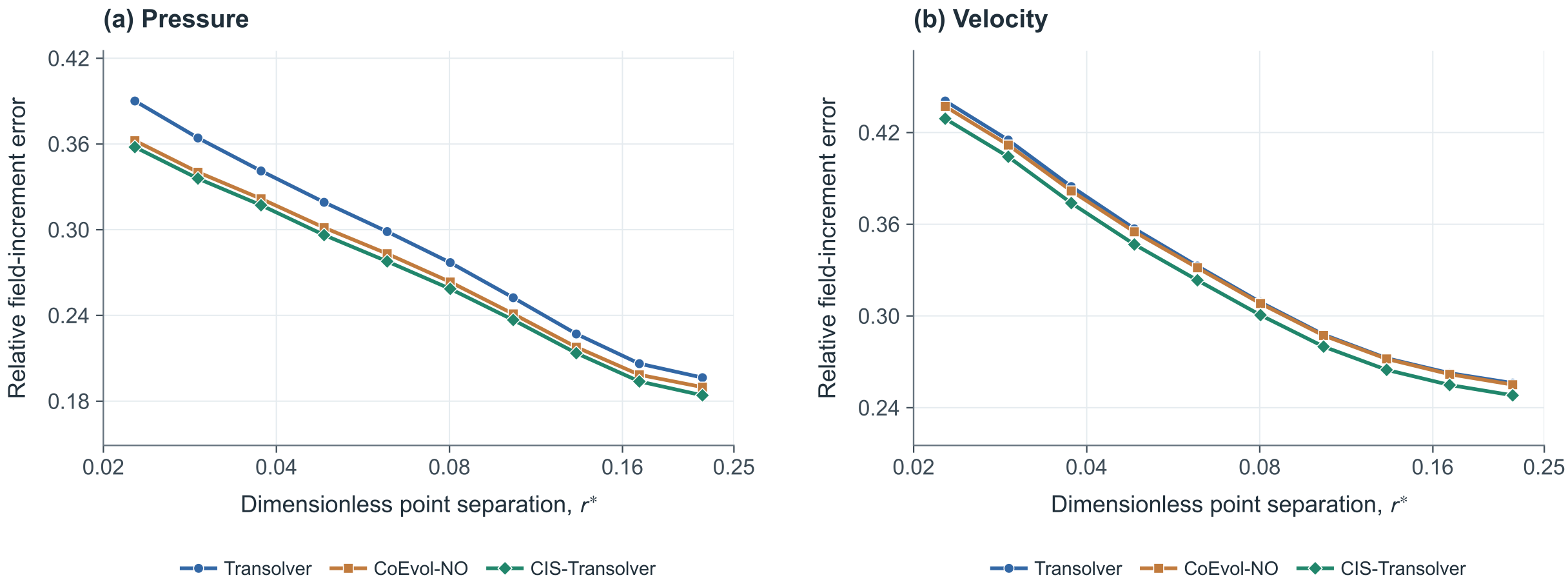

Figure 6 Relative field-increment errors for pressure (a) and three-component velocity (b) on DrivAerNet++. Point separation is normalized by vehicle length, $r^{*}=r/L$. Markers denote geometric centers of ten logarithmic bins. Results are averaged equally over the ten test geometries.

## 4.3 Spatial spectral analysis

For each of the 10 DrivAerNet++ test cases, we apply a density-compensated direct nonuniform Fourier transform to pressure and velocity at the same 32,000 points after inverse standardization and weighted mean removal. We evaluate $k=fL$ from 1 to 24 along 16 Fibonacci-sphere directions, where $f$ is spatial frequency and $L$ is the streamwise vehicle length. Power spectra are normalized by the total reference power.

Relative spectral error is the RMS magnitude of complex residual Fourier coefficients divided by the reference RMS magnitude across the 16 directions. Velocity errors include all three components. Results are averaged equally across cases.

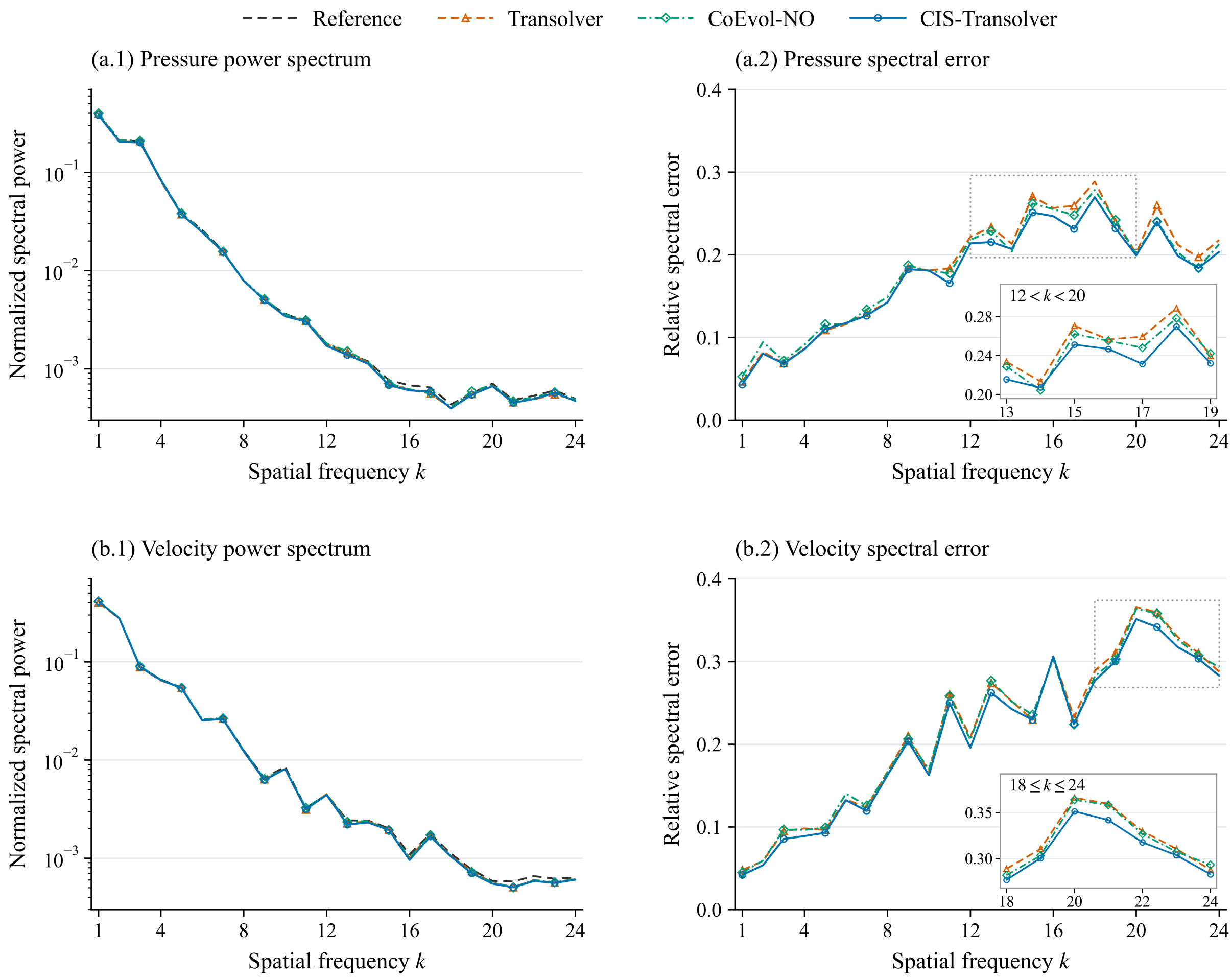

Figure 7 Normalized power spectra and relative spectral errors for pressure (top) and velocity (bottom). Insets enlarge higher-frequency intervals.

All three models reproduce the reference spectral decay in Figure 7, while the insets highlight higher-frequency differences. Over $13 \le k \le 24$, CIS-Transolver achieves mean pressure and velocity spectral errors of 0.2232 and 0.2865. These are 6.02% and 3.04% lower than those of Transolver, and 2.95% and 2.45% lower than those of CoEvol-NO, respectively.

Directional averaging in Figure 7 does not retain the angular distribution of residuals. Figure 8 therefore examines the $k_z = 0$ Fourier plane, where velocity residuals are more elongated along $k_y$ than pressure residuals. Within the high-frequency annulus $13 \le |k| \le 24$, CIS-Transolver has the lowest mean normalized residual amplitudes for both fields, with reductions of 5.24% for pressure and 3.05% for velocity relative to Transolver. These results indicate improved reconstruction of high-frequency variations in this plane.

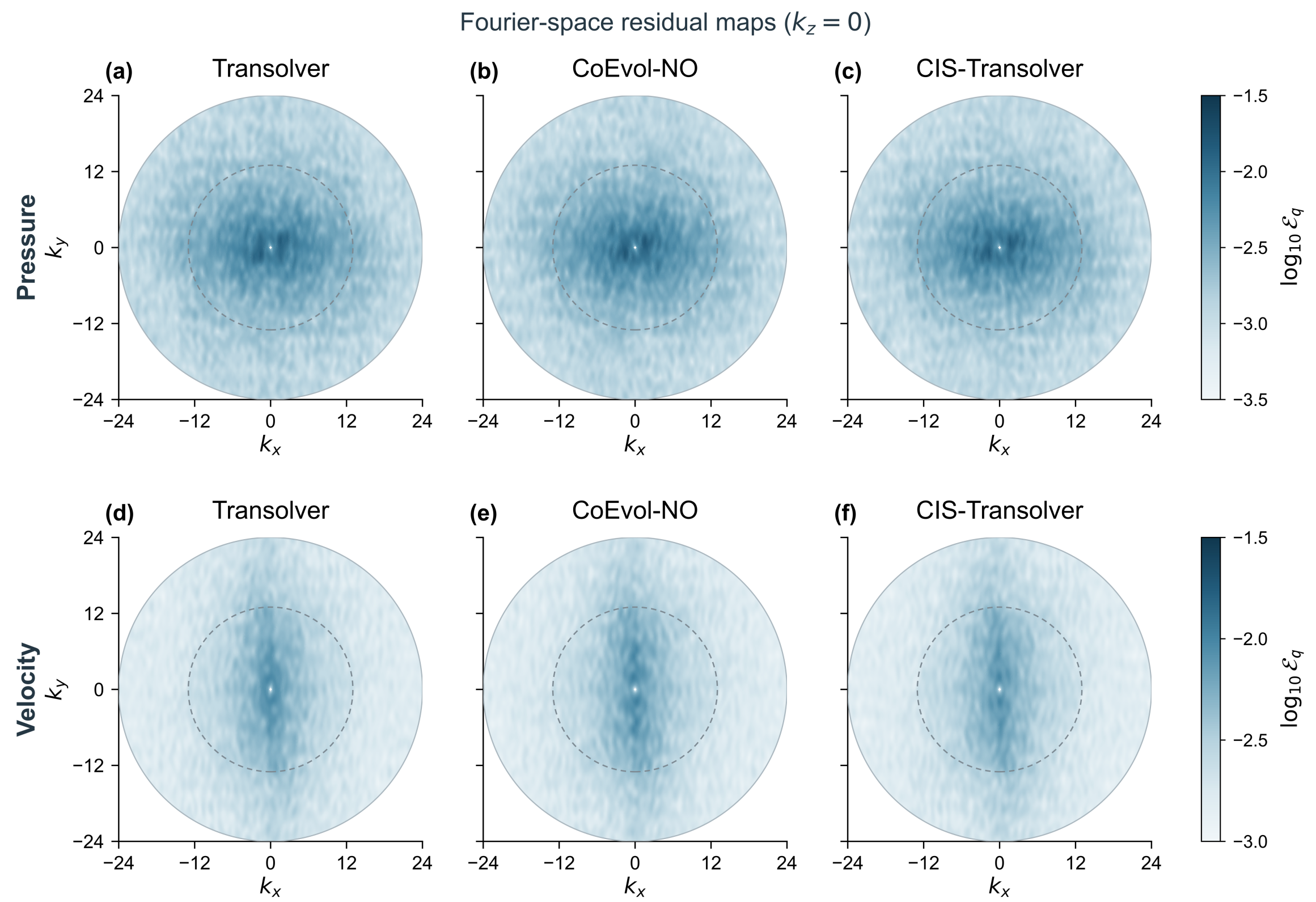

Figure 8 Fourier residual maps at $k_z = 0$: pressure (a–c) and three-component velocity (d–f). Columns: Transolver, CoEvol-NO, and CIS-Transolver. Grid spacing: 0.25. Amplitudes are normalized by the weighted reference fluctuation RMS, averaged equally over 10 cases, and displayed on a base-10 logarithmic scale. Color scales are shared within rows; dashed circles mark $|k| = 13$.

## 4.4 Ablation and capacity analysis

Component ablation and model-capacity comparisons examine the overall contribution of the CIS branch, the effect of cross-layer state reuse, and the relationship between performance gains and parameter count. All comparisons use the same 10 DrivAerNet++ test cases. Errors are computed for each case as defined in Section 4.1 and then averaged with equal weight across cases.

**Component ablation.** To assess the effectiveness of the CIS mechanism beyond the geometry-aware input representation, we include the original Transolver as a reference and consider two ablated variants. The first uses the Transolver backbone with the same geometry-aware inputs as CIS-Transolver but without the CIS branch, corresponding to the + Geometry encoding variant in Table 3. The second retains local detail extraction and current-layer Value feedback but removes cross-layer memory fusion and reuse, corresponding to the + Geometry encoding, Value feedback variant. This allows us to assess the additional benefit of using information across layers. Both variants are trained independently.

Table 3 Component ablation on DrivAerNet++

| **Method** | $\mathtt{rel}_p\downarrow$ | $\mathtt{rel}_U\downarrow$ | Spectral $p\downarrow$ | Spectral $u\downarrow$ |
|---|---|---|---|---|
| Transolver | 0.1501 | 0.1428 | 0.3149 | 0.2743 |
| + Geometry encoding | 0.1459 | 0.1445 | 0.3117 | 0.2709 |
| + Geometry encoding, Value feedback | 0.1420 | 0.1397 | 0.3054 | 0.2680 |
| **CIS-Transolver(Ours)** | **0.1405** | **0.1385** | **0.2946** | **0.2649** |

Note: Spectral errors are evaluated on 4,096 points per case. The high-frequency error is computed per case as the square root of the ratio of summed error power to summed reference power over $13 \leq k \leq 24$, then averaged equally across the ten cases.

Table 3 compares CIS-Transolver with the original Transolver and two ablated variants. The original Transolver serves as a reference, while CIS-Transolver and the two ablated variants share the same geometry-aware input representation. Compared with the + Geometry encoding variant, CIS-Transolver reduces the domain-wide pressure and velocity errors from 0.1459 and 0.1445 to 0.1405 and 0.1385, respectively, while the high-frequency pressure and velocity spectral errors decrease by approximately 5.5% and 2.2%. These results indicate that the complete CIS design provides additional improvements in overall prediction accuracy and high-frequency reconstruction beyond those achieved with the shared geometry-aware input representation.

The benefit of cross-layer state reuse is particularly evident in high-frequency pressure reconstruction. With local multiscale observations, moment-based state construction, and current-layer Value feedback retained, introducing cross-layer state reuse reduces the high-frequency pressure spectral error from 0.3054 to 0.2946, a relative reduction of 3.54%, compared with a 1.06% reduction in domain-wide pressure error. The high-frequency velocity spectral error also decreases from 0.2680 to 0.2649. Figure 9 examines how these improvements vary with spatial frequency.

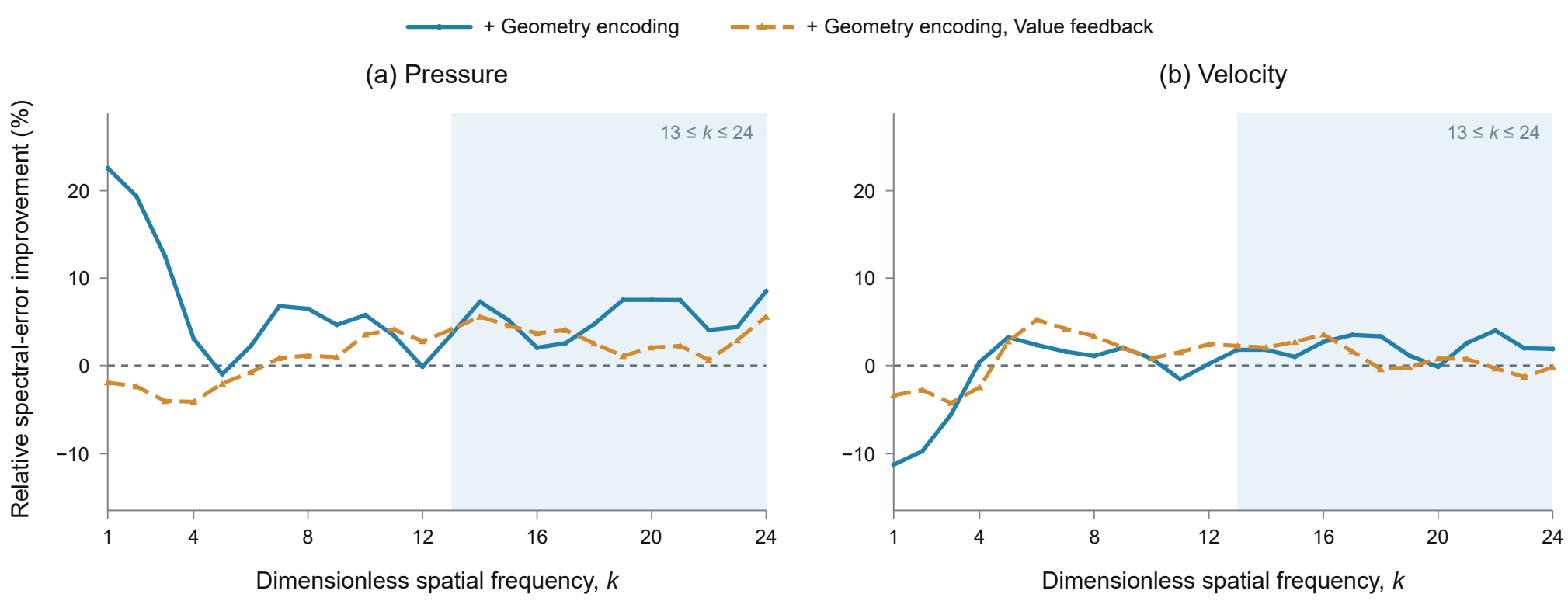

Figure 9 Frequency-resolved spectral-error reduction in the CIS ablations. (a) Pressure; (b) three-component velocity. Curves show the relative reduction in the ten-case mean spectral error of CIS-Transolver against each indicated ablation, using 4,096 points per case. Positive values indicate lower error for CIS-Transolver. Shading marks $13 \leq k \leq 24$.

Compared with the variant without the CIS branch, CIS-Transolver lowers pressure spectral errors over most of the evaluated frequency range, including the entire high-frequency interval (Figure 9a). Cross-layer state reuse further reduces pressure errors throughout $13 \leq k \leq 24$, whereas its effect on velocity varies within this interval (Figure 9b). These frequency-resolved results identify high-frequency pressure reconstruction as a particular benefit of retaining fine-scale states for subsequent layers, beyond local analysis and current-layer Value feedback.

**Model-capacity comparison.** We further compare CIS-Transolver with a wider Transolver variant, denoted Transolver-L, to assess their predictive performance at comparable parameter counts. Transolver-L contains 4.361M parameters, compared with 4.320M for CIS-Transolver.

Table 4 shows that Transolver-L achieves relative pressure and velocity errors of 0.1488 and 0.1465, respectively, whereas CIS-Transolver reduces these errors to 0.1405 and 0.1385. With slightly fewer parameters, CIS-Transolver achieves relative error reductions of 5.58% and 5.46%, respectively, compared with Transolver-L. These results demonstrate that the complete CIS-Transolver framework achieves better prediction accuracy than a higher-capacity Transolver baseline at a comparable model size.

Table 4 Model-capacity comparison on DrivAerNet++

| **Model** | **Parameters(M)** | $rel_p$ ↓ | $rel_U$ ↓ |
|---|---|---|---|
| Transolver | 3.865 | 0.1501 | 0.1428 |
| Transolver-L | 4.361 | 0.1488 | 0.1465 |
| CIS-Transolver | 4.320 | **0.1405** | **0.1385** |

## 4.5 Computational cost

Figure 10 shows that higher computational cost does not consistently translate into lower prediction error across the evaluated models, whose operation counts range from 59.8 to 1046.5 GFLOPs. CIS-Transolver achieves the lowest mean relative $L_2$ errors for pressure and velocity, at 0.1405 and 0.1385, respectively, with an intermediate forward cost of 295.0 GFLOPs. Lower-cost models yield larger errors in both fields, while GINO, AeroGTO, AB-UPT, and MeshGraphNet require more computation without matching this accuracy.

Relative to Transolver, CIS-Transolver requires 13.8% more forward computation and reduces the pressure and velocity errors by 6.3% and 3.0%, respectively. These improvements extend the accuracy of the backbone model to both predicted fields with a modest increase in operation count.

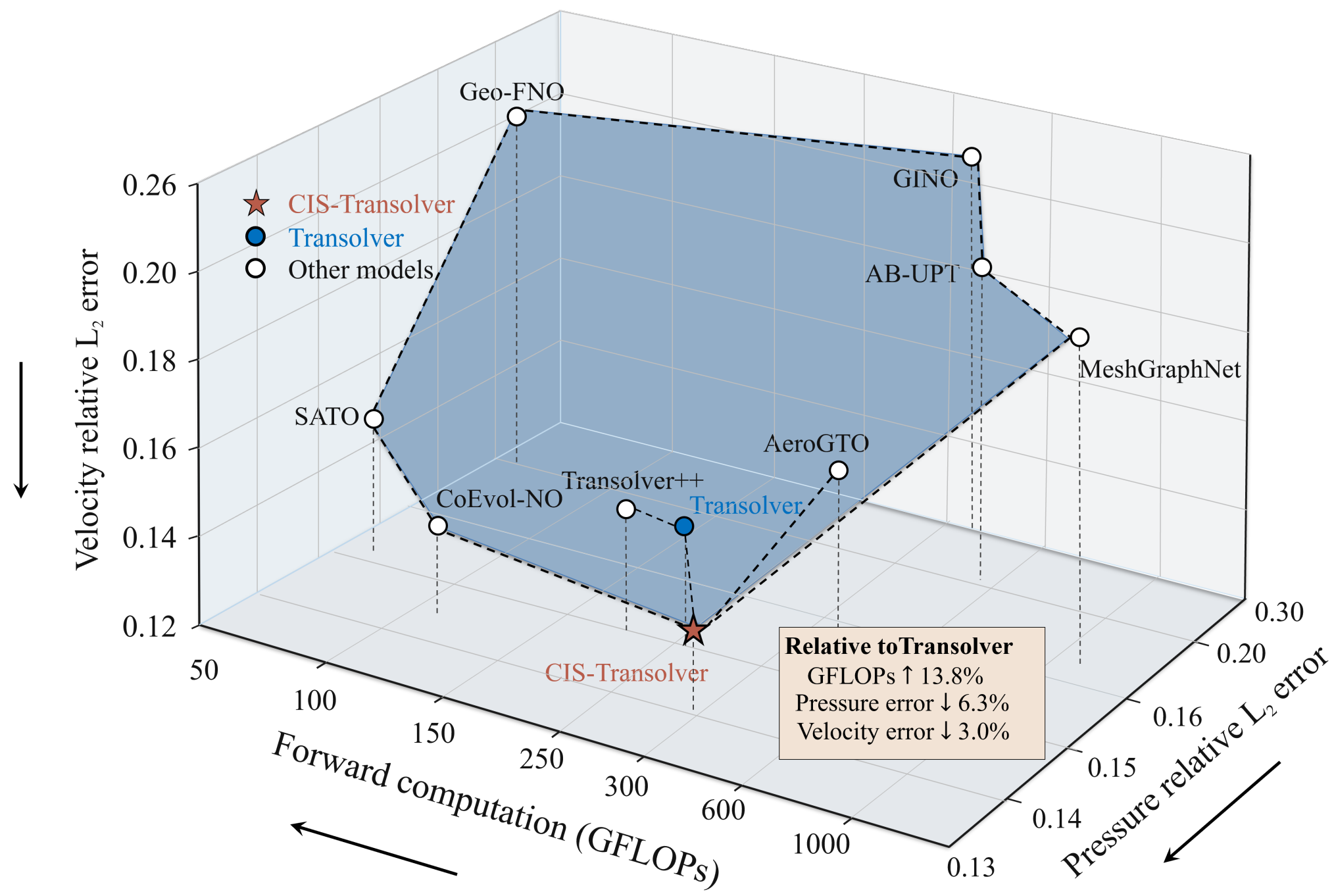

Figure 10 Prediction accuracy versus forward computational cost on DrivAerNet++. The green star denotes CIS-Transolver, the yellow circle denotes Transolver, and white circles denote the other models. GFLOPs are estimated for a single forward pass with 32,000 volume sampling points, a batch size of 1, and BF16 mixed precision. One multiply-add counts as two floating-point operations. Data loading, input preprocessing, and initial geometry cache construction are excluded. The shaded surface and dashed lines serve as visual guides.

## Conclusion

Accurate reconstruction of fine-scale flow structures around complex vehicle geometries remains challenging, particularly in capturing local spatial variations while accounting for the overall flow distribution. To address this challenge, we propose CIS-Transolver, a cross-layer incremental-state neural operator that combines local multiscale analysis with geometric moment memory. The model encodes local feature variations as zeroth- to second-order geometric moments and updates the resulting memory across operator layers. Memory-derived corrections enter the value branch of Physics-Attention, enabling extracted local information to participate in subsequent global operator updates and contribute throughout flow-field reconstruction.

Across ShapeNet-Car, AhmedML and DrivAerNet++, CIS-Transolver achieves the lowest mean relative errors on all five prediction metrics, with an average reduction of 8.72% relative to the best baseline for each metric. On DrivAerNet++, pressure and velocity increment errors at the shortest evaluated separation are 8.3% and 2.6% lower than those of Transolver, respectively, providing evidence of improved reconstruction of local spatial variations. Ablation experiments show that, with local analysis and feedback at refresh layers retained, cross-layer state reuse reduces domain-wide pressure error, with a larger relative reduction of 3.54% in high-frequency pressure spectral error. CIS-Transolver also achieves lower pressure and velocity errors than a wider backbone with a similar parameter count.

These results support the explicit organization and continued use of local information across layers as a component of fine-scale operator design. In the present model, gated fusion adapts memory content to new observations, while the same geometric neighborhoods are used across layers and refreshes occur at predetermined depths. However, as local representations evolve, the spatial scales and depths at which additional observations are most useful may also change. Further development will therefore focus on adaptive spatial support and memory-refresh timing to better align local observations with geometric complexity and evolving feature scales while preserving useful information from earlier layers.

## Appendix A. Supplementary experimental details and results

This appendix reports dataset preprocessing, evaluation protocols, supplementary flow-field comparisons, and state-refresh scheduling results.

### A.1 Experimental details

Datasets are split by complete geometry. Table A.1 lists the case counts and valid input points retained after preprocessing; AhmedML and DrivAerNet++ use fixed case subsets. Input features and encoders are specified in Appendix B.1.

AhmedML pressure is centered by its mean over valid fluid points; ShapeNet-Car and DrivAerNet++ retain the source pressure reference. AhmedML targets and ShapeNet-Car pressure are clipped channel-wise at the training-set 0.1st and 99.9th percentiles and standardized using the clipped training targets. DrivAerNet++ uses standardization without target clipping. All thresholds and standardization statistics are estimated from the training set and fixed for validation and testing. Evaluation reverses standardization to restore the units of the preprocessed targets.

Table A.1. Dataset splits and input point sets.

| Dataset | Train / validation / test | Input points per case | Target fields |
|---|---|---|---|
| ShapeNet-Car | 637 / 50 / 111 | 3,586 | Surface pressure |
| AhmedML | 100 / 20 / 20 | 32,000 | Volumetric pressure and velocity |
| DrivAerNet++ | 180 / 10 / 10 | 32,000 | Volumetric pressure and velocity |

Training objectives, optimization, and checkpoint selection are given in Appendix B.2 and Table B.2.

Full-field pressure and three-component velocity errors use equal point weights on the valid points of each case and are averaged equally over test geometries (Section 4.1). These metrics are computed before any surface mapping or interpolation used for display.

All methods use the same test cases and retain their input encodings and training configurations. On DrivAerNet++, CIS-Transolver also uses local and global geometric descriptors, whereas the archived baselines use coordinates and distance information. Section 4.4 examines the CIS branch through controlled ablations.

## A.2 Additional flow-field comparisons and error analyses

Figures A.1–A.5 compare four randomly selected test cases per dataset. Each row shows a reference field followed by the absolute errors of Transolver, CoEvol-NO, and CIS-Transolver. Pressure and velocity figures use the same four cases in the same row order. Viewpoints and error scales are shared across models within each row.

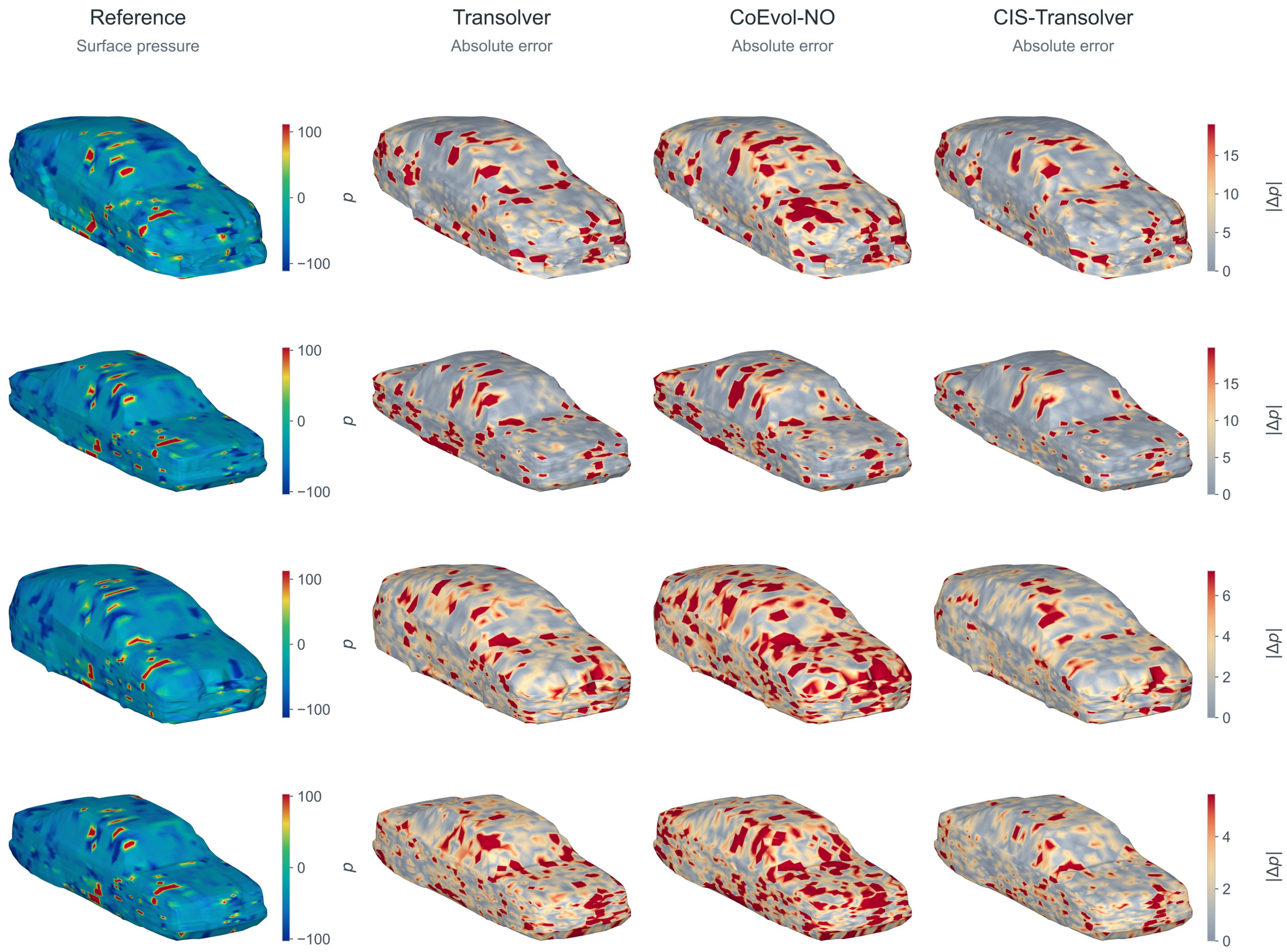

Fig. A.1. Surface pressure and absolute errors for four randomly selected ShapeNet-Car test cases. Each row represents a different test geometry, while the columns show the reference pressure field and the absolute errors of Transolver, CoEvol-NO, and CIS-Transolver.

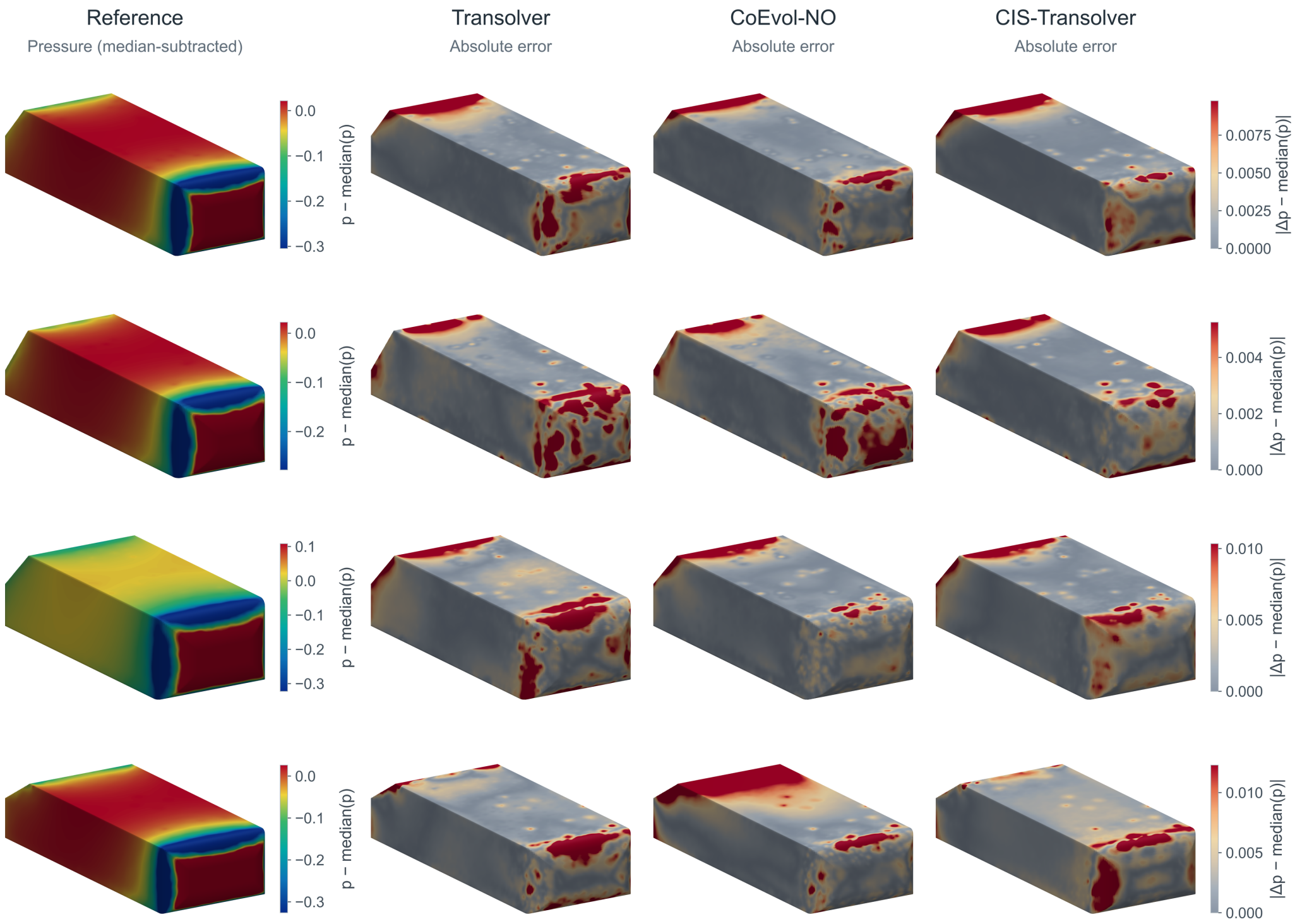

Fig. A.2. Pressure and absolute errors for four randomly selected AhmedML test cases. The volumetric field is mapped onto the geometric surface for display. Pressure is median-centered, and errors are computed in the same representation.

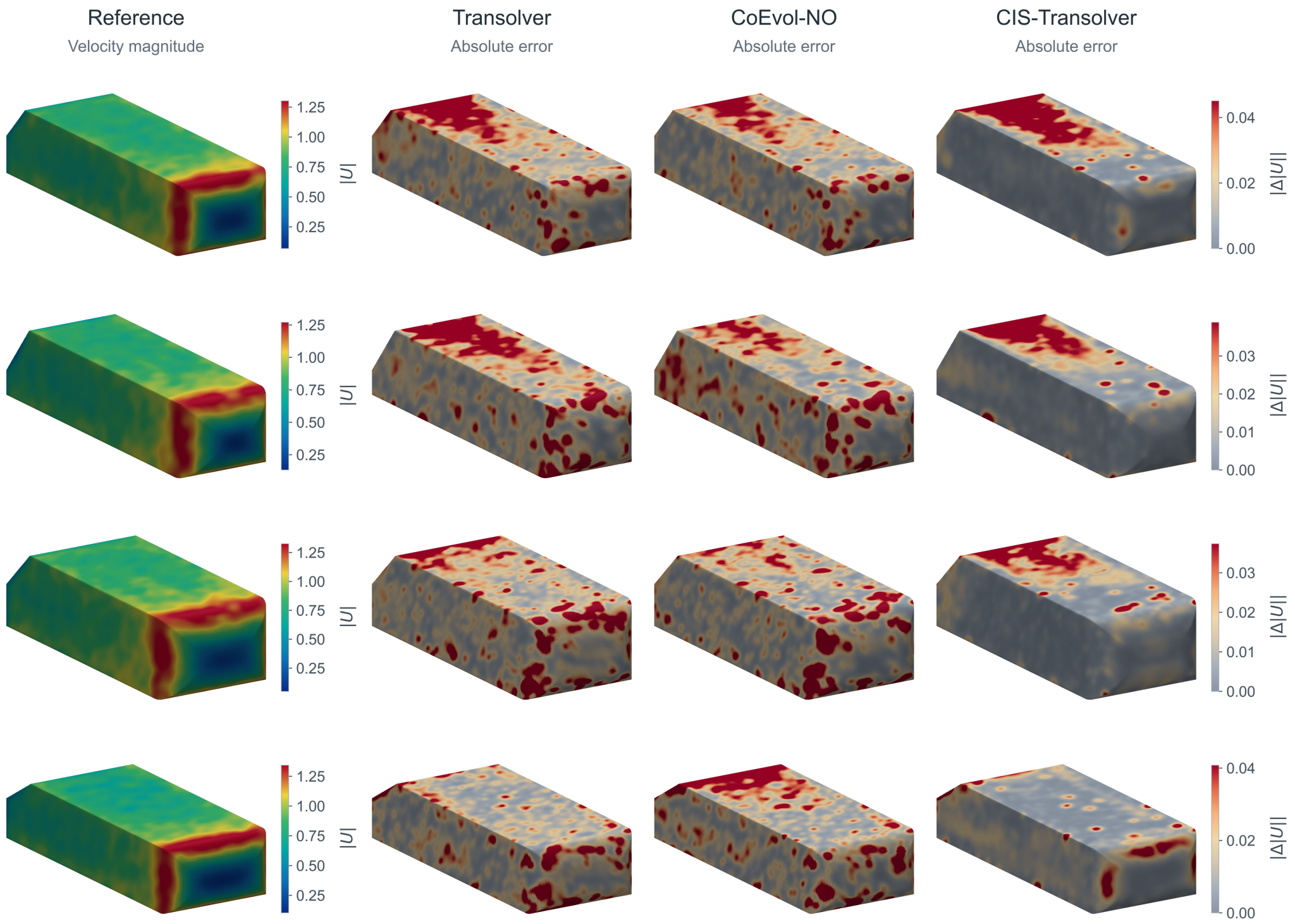

Fig. A.3. Velocity magnitude and its absolute error for the same four AhmedML test cases shown in Fig. A.2. Case order and column layout are unchanged. The volumetric field is mapped onto the geometric surface for display.

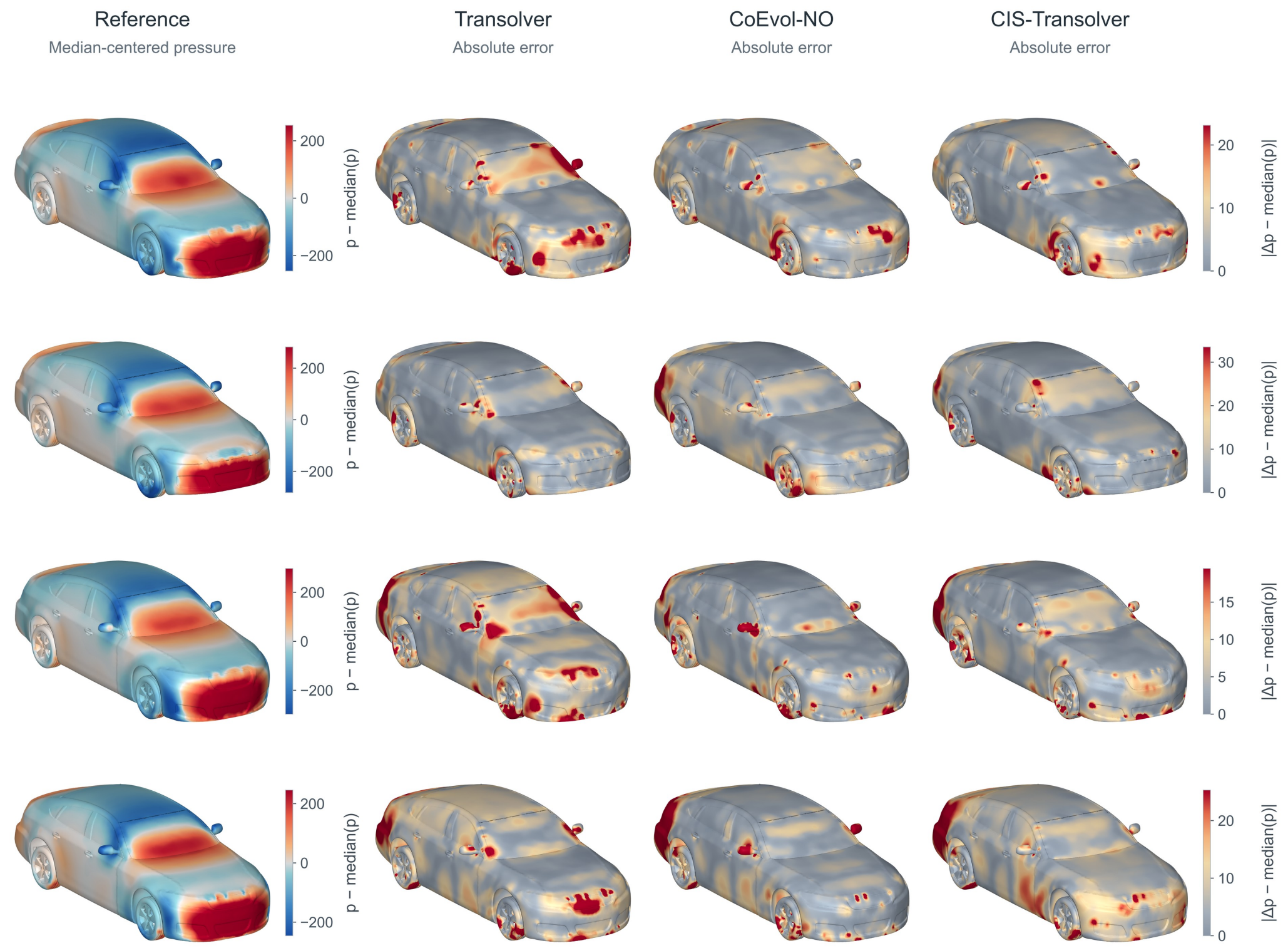

Fig. A.4. Pressure and absolute errors for four randomly selected DrivAerNet++ test cases. The volumetric field is mapped onto the geometric surface for display. Pressure is median-centered, and errors are computed in the same representation.

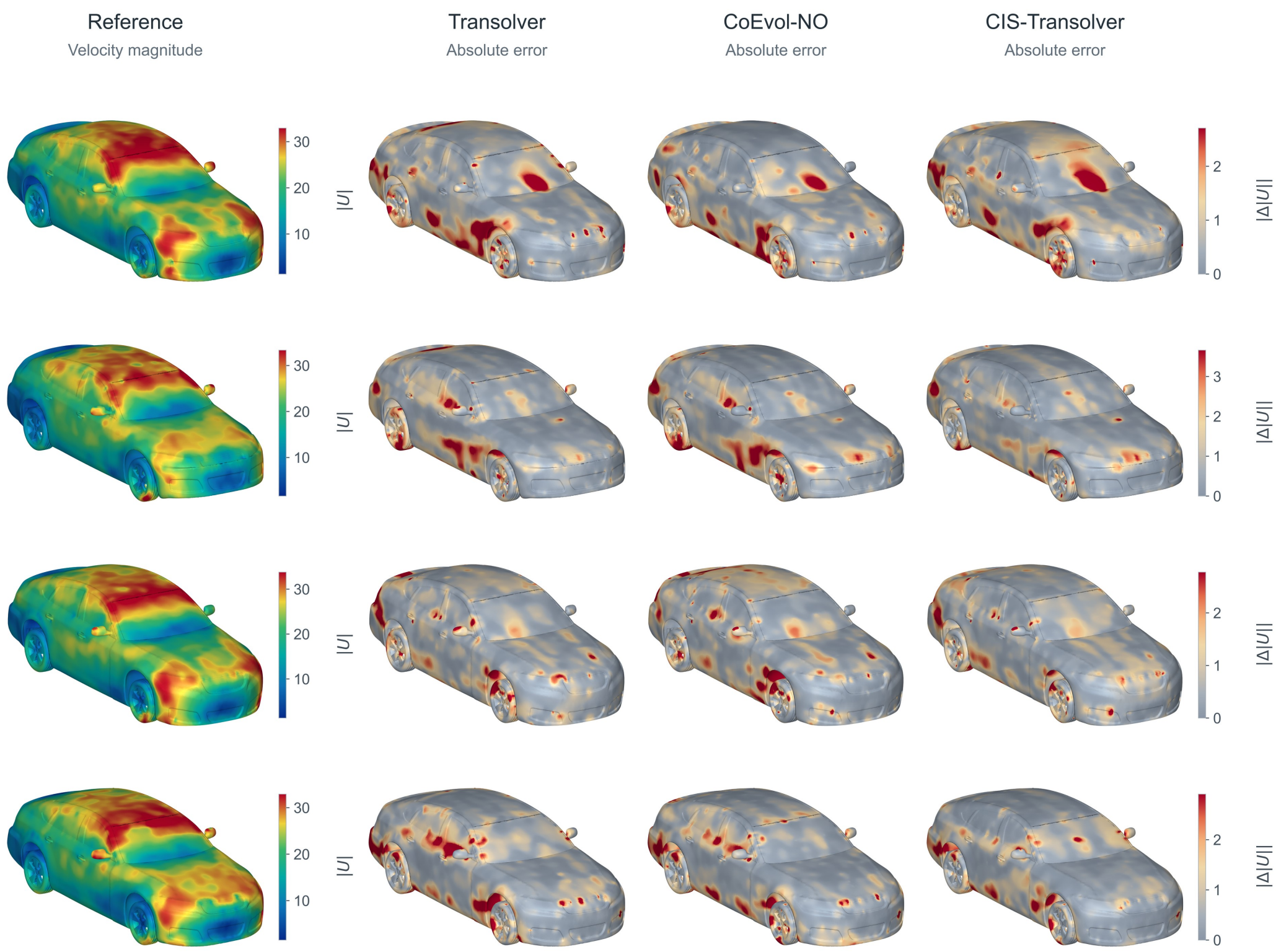

Fig. A.5. Velocity magnitude and its absolute error for the same four DrivAerNet++ test cases shown in Fig. A.4. Case order and column layout are unchanged. The volumetric field is mapped onto the geometric surface for display.

The ten shared DrivAerNet++ test geometries (0194–0203) are paired by identifier. We draw 20,000 bootstrap samples of ten cases with replacement, using identical sampled indices for all models. In each sample, the reduction is the difference between the mean baseline and CIS-Transolver errors divided by the mean baseline error. The 2.5th and 97.5th percentiles define the 95% confidence intervals in Table A.2; all four intervals are positive.

Table A.2. Paired case-level comparisons on DrivAerNet++.

| **Comparison model** | **Field** | **Comparison error** | **CIS error** | **Reduction (%)** | **95% CI (%)** |
|---|---|---|---|---|---|
| Transolver | Pressure | 0.1501 | **0.1405** | 6.34 | [4.85, 8.05] |
| Transolver | Velocity | 0.1428 | **0.1385** | 3.01 | [1.88, 4.31] |
| CoEvol-NO | Pressure | 0.1449 | **0.1405** | 2.97 | [0.39, 5.95] |
| CoEvol-NO | Velocity | 0.1421 | **0.1385** | 2.55 | [1.47, 3.70] |

Note: Means are rounded to four decimal places; reductions and intervals use unrounded case errors. Resampling uses PCG64 with seed 20261013. Intervals describe case-resampling variability at fixed checkpoints.

**Field-increment evaluation.**

For each DrivAerNet++ test case, we use all unordered pairs of the 32,000 valid volume points. Their Euclidean separation is normalized by streamwise vehicle length, $r^* = \frac{\| x_j - x_i \|}{L}$. For pressure or three-component velocity $q$, define $\delta q_{ij} = q_j - q_i$, with predicted increments defined analogously. In a separation interval $I$, the relative increment error is

$$E_{inc}(q, I = \sqrt{\frac{\sum_{(i,j) \in P_I} w_i w_j \| \delta \hat{q}_{ij} - \delta q_{ij} \|^2}{\sum_{(i,j) \in P_I} w_i w_j \| \delta q_{ij} \|^2}} \tag{A.1}$$

Here, $P_I = \{(i, j) : i < j, \frac{r_{ij}}{L} \in I$. Raw point weights $d_{16,i}{}^3$ use the distance to the 16th nearest neighbor, are clipped at their within-case 5th and 95th percentiles, and are normalized to sum to one. Models share the same points, pairs, and weights; per-case errors are averaged equally over the ten geometries.

Main-text Fig. 6 uses ten logarithmic bins over $0.02 < r^* \leq 0.25$. The lower limit exceeds the largest across-case weighted 90th-percentile nearest-neighbor spacing (approximately $0.015 L$); the upper limit defines a quarter-length separation window. Figure A.6 extends the analysis to $0 < r^* \leq 1$ using 40 equal-width bins and tests aggregation windows with the same lower limit and upper limits of $0.15\,L, 0.20\,L, 0.25\,L, 0.30\,L$. For each window, pairwise error and reference energies are pooled before taking the ratio and square root in Eq. (A.1).

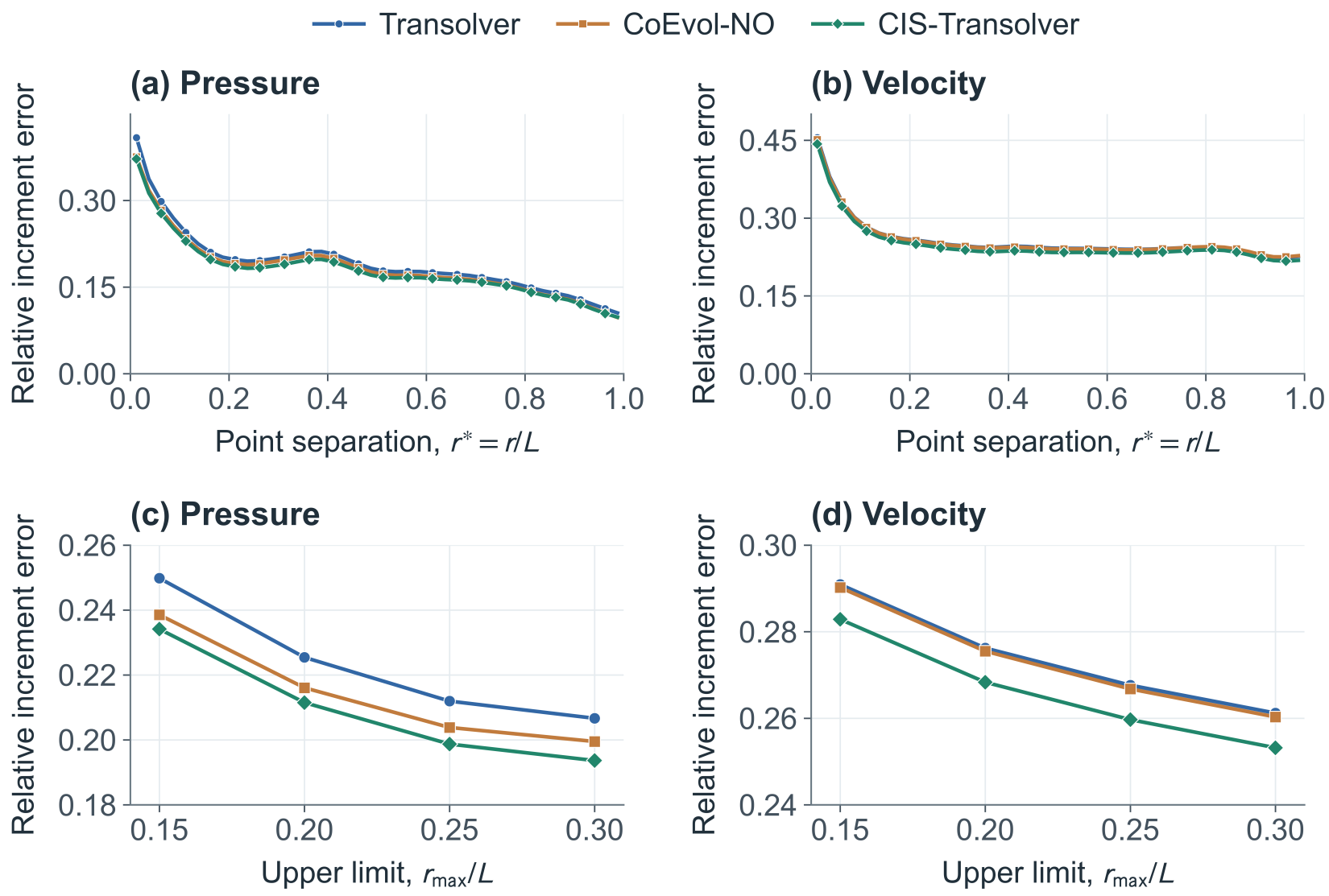

Fig. A.6. Field-increment errors on DrivAerNet++. (a, b) Pressure and three-component velocity errors over $0 < r^* \leq 1$ in 40 equal-width bins. (c, d) Errors pooled over $0.02 < r^* \leq \frac{r_{max}}{L}$. Results are computed per case and averaged equally over the ten test geometries.

CIS-Transolver retains the lowest window-aggregated pressure and velocity increment errors for all four upper limits, supporting the ranking across the tested separation windows.

## A.3 Sensitivity to state-refresh scheduling

We compare refreshes at layers 1 and 8, at layers 1, 4, and 8, and at all eight layers within a bounded-update variant. All schedules read memory through the Value path at every layer, with the first refresh initializing memory.

The variant caps the RMS of observation and history encodings at 2 per moment order and the RMS of the aggregated Value correction at 0.1 times the main-branch Value. The feedback gate is initialized to 0.1. Moment-content, detail-feedback, and feedback-constraint losses are applied at the final layer with weights 0.02, 0.02, and 0.01. All schedules use the same DrivAerNet++ splits, normalization, and training configuration: eight layers, 4.320M parameters, 520 epochs, and seed 0.

Table A.3. State-refresh scheduling in the bounded-update variant.

| Refresh layers | $\mathrm{rel}_p\downarrow$ | $\mathrm{rel}_U\downarrow$ | Spectral $p\downarrow$ | Spectral $u\downarrow$ |
|---|---|---|---|---|
| 1, 8 | 0.1432 | 0.1397 | 0.2295 | 0.2885 |
| 1, 4, 8 | **0.1415** | 0.1392 | **0.2280** | **0.2875** |
| 1–8 | 0.1428 | **0.1386** | 0.2284 | 0.2877 |

Note: Full-field errors use the original test records. Spectral errors follow Section 4.3, using 32,000 points per case and equal averaging over frequencies 13–24 and the ten cases. Boldface marks the lowest errors.

Refreshes at layers 1, 4, and 8 yield the lowest full-field pressure error and both high-frequency errors; refreshing every layer yields the lowest full-field velocity error.

## Appendix B. Model implementation and auxiliary objectives

This appendix supplements the operators in Section 3 with input configurations, memory readout, and training objectives.We use $P=P_g$, $R=R_g$, and $S=L_g$.

### B.1 Input encoding and geometric operators

The backbone has eight layers, width 256, eight heads, 32 physical states per head, and a 512-wide feed-forward block with GELU and pre-LayerNorm. Table B.1 lists the dataset-specific inputs and geometric operators; query coordinates are supplied separately.

Table B.1. Dataset-specific input interfaces and encoder configurations.

| Setting | ShapeNet-Car | AhmedML | DrivAerNet++ |
|---|---|---|---|
| Point attributes | Normal(3)<br>Condition (6) | SDF (1), fluid (1)<br>Condition (8) | SDF (1), surface (3)<br>Normal (3), direction (3)<br>Curvature (2), sharpness (1) |
| Attribute/condition dimension | 9 / 6 | 10 / 8 | 13 / 9 |
| Encoder output width | 256 | 256 | 256 |
| Anchor selection | Index-spaced | Index-spaced | Farthest-point |
| Interpolation weights | Inverse distance | Inverse distance | Local Gaussian |
| Fine reconstruction | $PS_aF$ | $PS_aF$ | $PRF$ |

Note: SDF denotes signed distance; surface denotes normalized nearest-surface-point coordinates; direction points from that surface point to the query. $S_a$ denotes direct anchor sampling and $R$ weighted aggregation. Index-spaced selection uses rounded, uniformly spaced indices in the input point ordering.

Case conditions encode vehicle dimensions and shape descriptors. ShapeNet-Car and AhmedML use training-set standardization; DrivAerNet++ normalizes lengths, areas, and volumes by vehicle height, squared height, and cubed height, respectively.

DrivAerNet++ surface coordinates map from $[0,1]^3$ to $[-1,1]^3$; normals and directions are unit vectors. Height-scaled curvatures use $asinh(clip(\kappa,-20,20))/asinh(20)$, and sharpness is clipped to $[0,1]$.

Query coordinates use the point-set bounding-box center and half its longest span for ShapeNet-Car and AhmedML. DrivAerNet++ uses the vehicle bounding-box center and half its longest span plus truncation distance. SDF is scaled by truncation distance for AhmedML and vehicle height for DrivAerNet++.

Flow-aligned coordinates use the same normalization for ShapeNet-Car and AhmedML; DrivAerNet++ uses the point-set mean and centered-coordinate RMS. The axial and transverse coordinates are $\xi,\eta_1,\eta_2$, with $\rho^2=\eta_1^2+\eta_2^2$. Normalized distance $d$ is SDF divided by its stop-gradient RMS and clipped to $[-8,8]$.

The learnable near-wall and wake bases are $a_d \exp\left(\frac{-|d|}{s_j}\right)$ and $\sigma\left(k_j(\xi-c_j)\right)\exp\left(\frac{-\rho^2}{r_j}\right)$ for ShapeNet-Car and AhmedML, and $\exp\left(\frac{-|d|}{s}\right)$ and $\sigma\left(\frac{\xi-c}{t}\right)\exp\left(\frac{-\rho^2}{r^2}\right)$ for DrivAerNet++. Scales are positive and centers bounded; $a_d$ disables unavailable distance information. ShapeNet-Car omits the near-wall basis.

Weights are proportional to $\left(d_{ia}+10^{-4}\right)^{-1}$ for ShapeNet-Car and AhmedML and $\exp[-(d_{ia}/s_g)^2]$ for DrivAerNet++, where $d_{ia}$ is the point–anchor distance and $s_g$ is the maximum distance over selected connections within each sample, floored at $10^{-3}$. Geometry matrices are constructed in FP32. Anchor masses are floored at $10^{-6}$ when forming $R$.

At refresh layers, hidden features are layer-normalized over all channels, split into attention heads, and projected to $r=8$ channels per head using bias-free, head-specific linear maps. Table B.1 uses $F$ for $F^l$ and lists the dataset-specific forms of $L_g(F^l)$.

## B.2 Moment-memory implementation and training

Inflow-aligned moment coordinates are centered at the point-set mean and divided by the maximum absolute displacement along each axis, floored at $10^{-4}$, to give $\overline{x}_i$. Anchor-wise centers and scales are

$$\mu_a = \sum_i R_{ai} \overline{x}_i ,$$
$$\sigma_{aj} = \sqrt{max\left(\sum_i R_{ai} \overline{x}_{ij}^2 - \mu_{aj}^2, 0.01\right)}, \xi_{ia,j} = \frac{\overline{x}_{ij} - \mu_{aj}}{\sigma_{aj}} \quad \text{(B.1)}$$

Moment encoding uses $r=8$ channels per order. Quadratic terms follow $xx, xy, xz, yy, yz, zz$ and are centered by their weighted means. Update maps are shared across anchors and scales and independent across layers and heads.

Observation and history maps are identity-initialized for the zeroth order and all orders, respectively. Update and readout gates start at 0.7 and zero.

The refresh schedule follows Section 3.3; Appendix A.3 evaluates the bounded-update variant.

For each head, $h_i^l$ is the corresponding slice of $H^{l-1}$, projected to $z_i^l$ with $r$ channels. A linear map of the concatenated scale memories produces anchor modulation and bias; $P$ interpolates them to $a_i^l, b_i^l$. At refresh layers, the bias is first filtered by $I - RP$. The readout is

$$q_i^l = \frac{1}{2} \tanh\left(a_i^l\right) \odot z_i^l + \frac{1}{2} b_i^l ,$$
$$u_i^l = \Phi_{out}^l\left(q_i^l\right), \Delta v_i^l = \gamma_l g_i^l s_i^l u_i^l . \quad \text{(B.2)}$$

A learned affine map takes three detached ratios: the channel RMS of $z_i^l$, $q_i^l$, and $q_i^l - z_i^l$, each divided by the channel RMS of $h_i^l$. Its tanh output is $g_i^l$. Amplitude matching and the layer gain are

$$s_i^l = sg\left(clip\left(\frac{RMS\left(h_i^l\right)}{RMS\left(u_i^l\right)}, \frac{1}{4}, 4\right)\right),$$
$$\gamma_l = 0.72\, \sigma(b)\, softmax(\beta)_l . \quad \text{(B.3)}$$

RMS is over channels, with a $10^{-6}$ floor on the mean square; $sg$ stops gradients. $b$ and $\beta$ are learned budget and depth-allocation parameters. $\Delta V^l$ corresponds to $\Delta^l$ in the main text.

State assignments use a softmax over states with temperature $softplus(t_l) + 0.05$, initialized to 0.7. State masses are floored at $10^{-6}$. The same mass-normalized weights pool point values and memory corrections; the correction is added after the head-specific Value projection (main-text Eq.

(8)). State attention includes the shortcut $\chi S^l$, with $\chi=0$ for ShapeNet-Car and AhmedML and $\chi=1$ for DrivAerNet++.

**Training objective, optimization, and model selection**

Training follows the two-term objective in Section 3.4. Both losses use the final prediction and reference field in standardized target units. For volumetric tasks, pressure MSE and the mean MSE over the three velocity components receive equal weight; surface tasks use pressure alone. Invalid entries are zeroed before filtering and excluded from averaging.

The detail operator $H_g = I - PR$ reuses the model geometry and filters the final prediction error without detaching it. The separate physical-moment probe is frozen and used only for diagnostics.

Field loss is evaluated every training forward pass; detail loss is evaluated every $k$ passes and multiplied by $k$, with zero contribution otherwise (Table B.2). Both terms are evaluated on every validation batch without interval scaling. Diagnostic requests leave the training schedule unchanged.

Table B.2. CIS-Transolver training configuration for the main experiments.

| Setting | ShapeNet-Car | AhmedML | DrivAerNet++ |
|---|---|---|---|
| Training epochs | 900 | 900 | 520 |
| Cosine schedule (epochs) | 900 | 900 | 460 |
| Warmup fraction | 4% | 4% | 3.5% |
| AdamW betas | (0.9, 0.999) | (0.9, 0.999) | (0.9, 0.95) |
| Gradient-norm limit | 0.8 | 0.8 | 1.0 |
| Field-loss coefficient | 1.00 | 1.00 | 1.00 |
| Velocity / pressure weights | Pressure only | 0.50 / 0.50 | 0.50 / 0.50 |
| Detail interval (training) | 4 | 4 | 8 |
| Detail interval (validation) | 1 | 1 | 1 |

Note: Intervals count forward passes. Warmup is relative to schedule length. DrivAerNet++ remains at the minimum learning rate after epoch 460. The bounded-update variant uses the separate protocol in Appendix A.3.

All tasks use AdamW, batch size 2, one gradient-accumulation step, and seed 0, with initial learning rate $1.8\times10^{-4}$, minimum rate $5\times10^{-7}$, and weight decay $8\times10^{-5}$. Training uses linear warmup, cosine decay, BF16, and TF32. Table B.2 gives the dataset-specific settings.

Validation relative-L2 errors determine checkpoint selection. ShapeNet-Car minimizes mean pressure error and AhmedML the mean of pressure and velocity errors. DrivAerNet++ uses the score

$$s = max(e_U, \ e_p) + 0.10\sqrt{\frac{e_U^2 + e_p^2}{2}} + 0.05 e_{all} \tag{B.4}$$

Here, $e_U$ and $e_p$ are validation-case mean errors. For $e_{all}$, inverse-standardized velocity and pressure are concatenated before computing case errors and averaging. The selected checkpoint is evaluated on the test set.